\documentclass[10pt,twocolumn]{article}

\usepackage[T1]{fontenc}
\usepackage[utf8]{inputenc}
\usepackage[margin=1.9cm,columnsep=0.6cm]{geometry}
\usepackage{times}

\usepackage{amssymb}
\usepackage{amsmath}

\usepackage{graphicx}
\usepackage{booktabs,subcaption}
\usepackage{multirow}

\usepackage[colorlinks=true,linkcolor=RoyalBlue,citecolor=RoyalBlue,urlcolor=RoyalBlue]{hyperref}
\usepackage{comment}

\usepackage{cleveref}
\usepackage{csquotes}
\usepackage{placeins}
\usepackage{stfloats}


\usepackage{enumitem}

\usepackage[dvipsnames]{xcolor}

\definecolor{pltorange}{HTML}{ff7f0e}
\definecolor{pltblue}{HTML}{1f77b4}

\definecolor{color1}{HTML}{8172b3}
\definecolor{color2}{HTML}{8c8c8c}
\definecolor{color3}{HTML}{7ea6e0}
\definecolor{color4}{HTML}{c44e52}
\definecolor{color5}{HTML}{da8bc3}
\definecolor{color6}{HTML}{ccb974}

\newcommand{\newtext}[1]{#1}
\newcommand{\newtextrm}[1]{}
\newcommand{\torm}[1]{}

\usepackage[numbers,sort&compress]{natbib}

\usepackage{authblk}

\title{\bf Precision at Scale: Domain-Specific Datasets On-Demand}

\author[1,2]{Jesús M. Rodríguez-de-Vera$^{*}$}
\author[1,2]{Imanol G. Estepa$^{*}$}
\author[3]{Ignacio Sarasúa}
\author[4]{Bhalaji Nagarajan}
\author[1,5]{Petia Radeva}

\affil[1]{Dept. de Matemàtiques i Informàtica, Universitat de Barcelona, Barcelona, Spain}
\affil[2]{Computer Vision Center, Cerdanyola del Vallès, Barcelona, Spain}
\affil[3]{NVIDIA Computing Spain, Madrid, Spain}
\affil[4]{Barcelona Supercomputing Center (BSC), Barcelona, Spain}
\affil[5]{Institut de Neurosciències, Universitat de Barcelona, Barcelona, Spain}

\date{}

\begin{document}

\twocolumn[
\maketitle
\vspace{-2.6em}

\begin{center}
\small $^{*}$Equal contribution. Corresponding author: Jesús M. Rodríguez-de-Vera (\texttt{j.molina.rdv@ub.edu}).\\
Code: \url{https://github.com/jesusmolrdv/Precision-at-Scale}\\
Data \& Models: \url{https://huggingface.co/jesusmolrdv}
\end{center}

\begin{center}
\fbox{\parbox{0.94\linewidth}{\small
This is the accepted manuscript of the article published as: J.~M. Rodríguez-de-Vera, I.~G. Estepa, I.~Sarasúa, B.~Nagarajan, P.~Radeva, \emph{Precision at scale: Domain-specific datasets on-demand}, Pattern Recognition 171 (2026) 112236. \url{https://doi.org/10.1016/j.patcog.2025.112236}.
\copyright~2026. This manuscript is made available under the CC-BY-NC-ND 4.0 license \url{https://creativecommons.org/licenses/by-nc-nd/4.0/}.
}}
\end{center}
\vspace{0.5em}

\begin{abstract}

Recent self-supervised learning methods rely on massive general-domain datasets for robust model pretraining. However, these datasets may lack specificity required in specialized domains. Collecting large, supervised datasets to compensate for this limitation is also cumbersome. This raises a key question: \emph{Can automatically crafted domain-specific datasets serve as efficient and effective SSL pretrainers, performing comparable to --- or even surpassing --- much larger state-of-the-art general-domain datasets?} To address this challenge, we propose \textbf{Precision at Scale (PaS)}, a novel modular pipeline for automatic creation of domain-specific datasets on-demand. PaS leverages Large Language Models (LLMs) and Vision-Language Models (VLMs) through three distinct phases: Concept Generation, where LLMs identify relevant domain concepts; Image Collection, utilizing VLMs and Generative models to gather appropriate images; Data Curation, ensuring quality and relevance by eliminating unrelated or redundant images. We conduct extensive experiments across three complex domains \textit{--- food, insects, and birds ---} proving that PaS datasets compete and often surpass existing domain-specific datasets in diversity, scale, and effectiveness as pretrainers. Models pretrained on PaS datasets outperform those trained on large-scale general-domain datasets (ImageNet-1K) by up to 21\% and surpass same-scale domain-specific datasets by 6.7\% across classification tasks. Notably, despite being an order of magnitude smaller, PaS datasets outperform ImageNet-21K pretraining, with improvements of 3.3\% in fine-tuning and 9.5\% in few-shot learning, and showing superior performance on specialized dense tasks. Furthermore, by efficiently fine-tuning pretrained VLMs like CLIP and SigLIP using low-rank methods, we achieve performance gains (+4.2\% over CLIP) in specialized domains with minimal overhead, demonstrating the versatility of PaS datasets.

\end{abstract}
\bigskip
]

\section{Introduction}

\textbf{Self-Supervised Learning (SSL)} models such as DINOv2 \cite{oquab2024dinov2}, trained on millions of images, obtain very high performance on most general discriminative tasks such as image classification and object detection \cite{gui2024survey, liu2021self}.
However, most SSL
methods focus on being as general as possible and require huge volumes of broad-domain images to ensure high performance in domain-specific tasks. 
These datasets,
refered as pretrainer datasets (in short, \enquote{pretrainers}),
are usually unsupervised or programmatically created.

Concurrently, various \textbf{domain-specific datasets} \cite{min2023large, bossard2014food, van2018inaturalist, jain2024insect} are designed to train
expert models in specific domains, trying to bridge the gap between existing computer vision solutions and real-world application in those domains.
Nevertheless, these supervised datasets require expensive investments in human annotations.
This leads to a small number of images compared to popular unsupervised datasets \cite{schuhmann2022laion, oquab2024dinov2} (For example, Food-2K \cite{min2023large}, the largest food benchmark has $\approx$600K training images, while LAION \cite{schuhmann2022laion} has 5.85B
images), making them suboptimal
for recent large models \cite{liu2021swin,dosovitskiy2020vit}. 
On the other hand, generic datasets like ImageNet lack comprehensive coverage for many specialized real-world domains.
{\em Domain-specific datasets, in general, prove to be better in their expertise, however, lack the scalability}
as the annotation process is not feasible at scale.
With recent generative models, synthetic datasets \cite{hammoud2024synthclip, tian2024learning} are gaining popularity. 
They show performance comparable to real ones and, importantly, provide scalability on demand.
However, they
fail to cover \enquote{under} represented (infrequent) classes.
SynCLR \cite{tian2024learning}, for example, 
leverages an initial set of general captions from supervised datasets to create a completely synthetic dataset,
making the dataset generation process
tightly coupled with \enquote{external supervision}. 
Ideally, a smart, well-balanced mix of unrestricted real and synthetic images could alleviate the problem.

\begin{figure}[!t]
    \centering
    \includegraphics[width=0.99\linewidth]{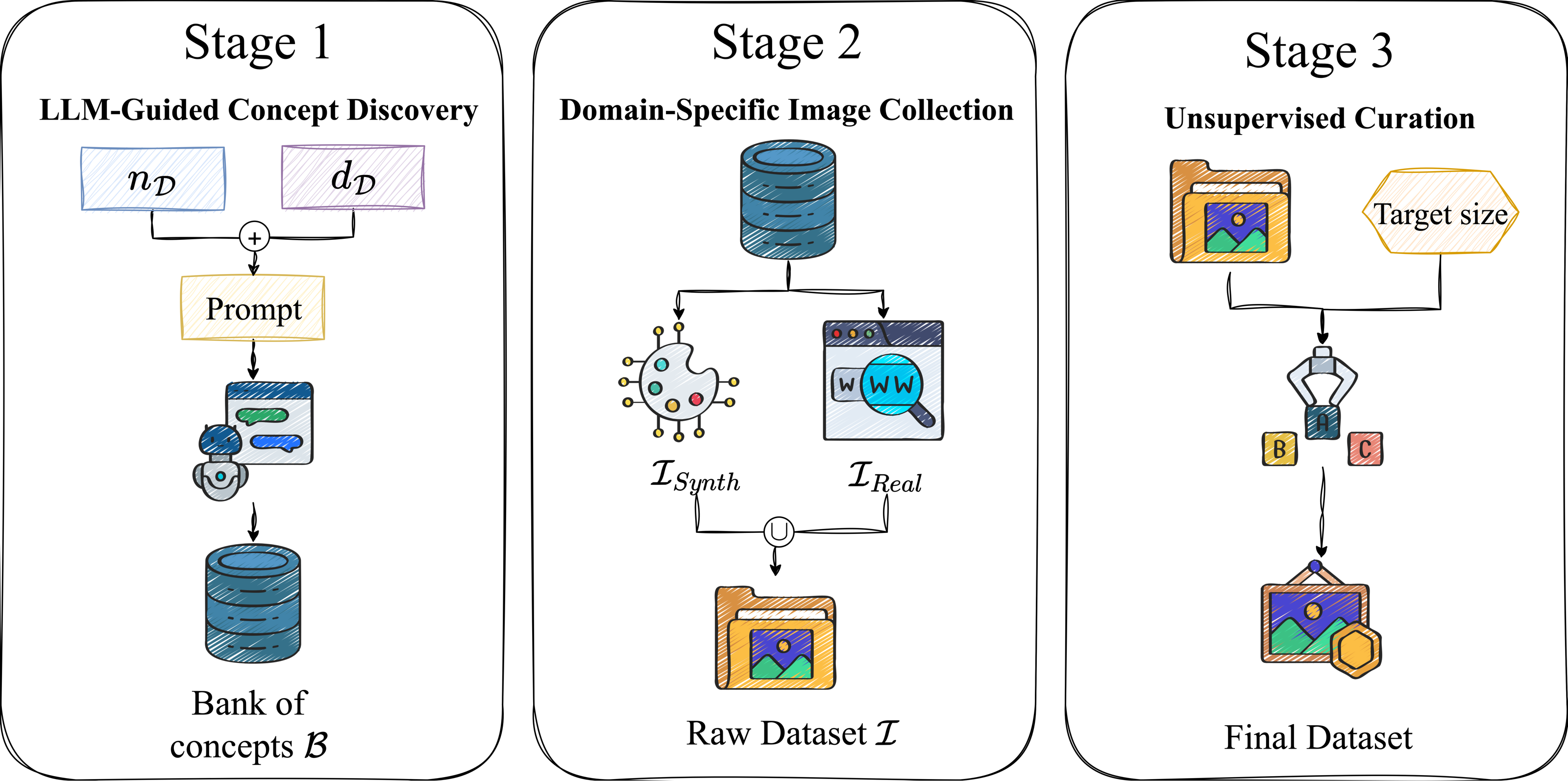}
\caption{
\textbf{PaS Pipeline Overview.}
With \enquote{only} two textual inputs describing a target domain ($n_\mathcal{D}$ and $d_\mathcal{D}$), PaS autonomously creates SoTA domain-specific datasets in 3 stages:
LLM-guided diverse concept bank creation (\textbf{Stage 1});
Web 
and synthetic 
images collection (\textbf{Stage 2}); 
Automatic removal of
OOD images, reducing dataset size while preserving
relevant information (\textbf{Stage 3}).}\label{fig:full-diagram}
\end{figure}

\textbf{
Two natural research questions arise:} 
\emph{(1) Is it possible to design a framework that automatically crafts domain-specific datasets, outperforming existing SoTA datasets? 
(2) Can these datasets serve as more efficient and effective pretrainers than much larger general-domain datasets, when the target domain is known?}
To address these questions, 
we introduce \textbf{Precision-at-Scale (PaS)} (illustrated in \Cref{fig:full-diagram}),
a general and modular pipeline to autonomously generate high-quality, domain-specific datasets on demand (\enquote{PaS datasets}), without any dependence on external labels and human experts.
PaS is
elegant - it leverages the synergy between LLMs and VLMs to create scalable, well-curated domain-specific datasets through 3 key stages:
(1) \textbf{LLM-guided Concept Discovery}
to generate a comprehensive, rich bank of domain-specific concepts,
(2) \textbf{Domain-Specific Image Collection} using diverse real and synthetic images,
(3) \textbf{Autonomous Dataset Curation}
using advanced unsupervised curation methods to
ensure high quality and domain relevance.
In addition, PaS offers an efficient task-agnostic analysis framework to assess the diversity of the generated PaS datasets \newtext{(see \Cref{sec:results-task-agnostic})}.

PaS is a novel framework for \textbf{efficient domain-specific learning}. PaS datasets perform better at a much smaller size (one order smaller), leading to less computation and faster training.
PaS is
flexible, making no assumptions on specific components, allowing integration of any LLM, VLM, or Image generation models.
This versatility ensures
seamless incorporation of newer models as they are developed.
PaS datasets match --- and even surpass --- the diversity of human-created domain-specific datasets \cite{min2023large, bossard2014food, wah2011caltech, Horn_2015_CVPR}, while providing more comprehensive coverage of real-world complex domains through a completely automated way.
\newtext{
Distinct from knowledge distillation, our framework does not simply transfer knowledge, but instead refines and structures it into a new, hybrid data asset. We demonstrate that this generates emergent value, as our PaS datasets can be used to fine-tune and improve the original foundation models (see \Cref{sec:results-vlm}).
}

\textbf{Extensive experiments} on multiple domains ---\textit{food}, \textit{birds} and \textit{insects}--- show
SSL models pretrained on PaS datasets not only outperform existing domain-specific SoTA datasets at the same scale, but also show superior performance to models trained on much larger general-domain datasets like ImageNet-1K (IN-1K) and even ImageNet-21K (IN-21K) on specific domain evaluation.
Beyond pretraining from scratch, PaS datasets offer a promising avenue for efficiently adapting existing VLMs to new domains. 
By leveraging automatically generated image-text pairs, we fine-tune a pretrained VLM on PaS datasets for a new domain, enabling cheap and on-demand adaptation. 
Remarkably, PaS datasets, through LoRA fine-tuning \cite{hu2022lora}, efficiently enhance the capabilities of pretrained VLMs models like CLIP \cite{radford2021learning} and SigLIP \cite{zhai2023sigmoid} in their respective domains at a tiny fraction of the computational and data cost required to train them.
The advantage of PaS datasets over traditional domain-specific datasets grows significantly when leveraging its scalability.
In summary, \textbf{our contributions} are:
\begin{itemize}[wide = 0pt]
\item  
We propose PaS, a novel \textbf{domain-specific dataset creation pipeline} (\Cref{sec:approach}), that, given a domain, creates a hybrid dataset of web and synthetic images, at a given scale and computational budget.
\item 
We define an \textbf{efficient low-resource task-agnostic diversity analysis framework} (\Cref{sec:results-task-agnostic}) and emphasize that PaS datasets provide better diversity than current SoTA domain-specific supervised datasets.
\item
With \textbf{extensive validation} (\Cref{sec:results-downstream}) on three complex domains,
we prove PaS datasets as much
better pretrainers, 
showing consistent superior performance across domains,
datasets and downstream tasks\newtext{, including classification, and dense tasks such as semantic segmentation and keypoint detection}.
Compared to general-domain datasets (IN-1K), PaS demonstrate that on the same and even smaller scale, domain-specific datasets outperform general ones, 
proving that quality stands over quantity on model pretraining.
\item
We demonstrate that PaS datasets effectively improve the performance of large pretrained VLMs in specific domains leading to substantial improvements (an average gain of 4.2\% for CLIP) with minimal computational overhead.

\end{itemize}

The rest of the paper is
as follows: \Cref{sec:related} discusses the recent related works.
Our proposed framework, Precision at Scale (PaS), is detailed in
\Cref{sec:approach}.
We then present a comprehensive,
diversity and domain alignment low-resource evaluation of PaS datasets in \Cref{sec:results-task-agnostic}. 
The effectiveness of PaS datasets as pretrainers and for fine-tuning is demonstrated in \Cref{sec:results-downstream}, followed by the conclusions
in \Cref{sec:conclusion}.\section{Related Works}\label{sec:related}

The increasing \textbf{demand for large-scale data} in training
models such as
ConvNeXt \cite{woo2023convnext} and
ViT \cite{zhai2022scaling}, 
makes dataset creation at scale very critical.
DINOv2 \cite{oquab2024dinov2} and Internet Explorer \cite{li2023internet} use 
sophisticated automatic data curation pipelines to curate high-quality real images.
Generative models like Stable Diffusion \cite{Rombach_2022_CVPR} and MUSE \cite{chang2023muse} have propelled the creation of synthetic datasets \cite{azizi2023synthetic}. 
SynCLR \cite{tian2024learning} 
uses labels of existing datasets, while
SynthCLIP \cite{hammoud2024synthclip} creates large-scale synthetic image-text pairs using
a predefined concept bank of MetaCLIP \cite{xu2024demystifying}. 
Despite having a vast collection of 500k concepts, the MetaCLIP concept repository fails to provide extensive coverage in certain specific domains 
(For example, MetaCLIP \enquote{only} contains only simple \textit{paella}, a popular Mediterranean food as a single concept without any of its common variations).
This shows that even large-scale concept banks may lack necessary domain specificity.
These methods can generate large datasets, however they rely on curated information, like predefined class labels or captions, limiting their adaptability to new domains. 
Their focus on scale can sometimes 
compromise
domain specificity.

Several works addressed \textbf{the need for domain-specific data.} 
Despite its noise, domain-specific web data proved more effective in fine-grained recognition \cite{krause2016unreasonable}. 
InfoGrowth \cite{Qin2024ECCV} and T-MARS \cite{maini2024tmars}
enabled dataset growth with maintained cleanliness and diversity.
Large unsupervised datasets like LAION-5B
warranted
extraction of custom subsets tailored for specific use cases \cite{li2023internet, zheng2022general}.
Alternatively, in linguistics, DoPAMine \cite{Arannil2024ARXIV}
mines domain-specific LLM training data.
CRAFT \cite{Ziegler2024ARXIV}
generates synthetic datasets by retrieving and augmenting corpora. 
Although focused on text,
these methods underscore the importance of domain-specific datasets. 
Recent studies have extended
this idea in vision.
Performance of SSL models prove to increase \enquote{only} when data aligns closely with the target domain \cite{Hammoud2024ECCV}.

\textbf{Self-Supervised Learning} enables learning adaptable features from unlabeled data \cite{chen2020simple, he2020momentum, dwibedi2021little}, reducing the reliance on extensive labeled datasets. 
Recent popular models like MoCoV3 \cite{chen2021empirical}, MAE \cite{he2022masked}, and CAE \cite{chen2024context}
have shown increased robustness and efficiency 
owing to
better computational power, model complexity, and data scale \cite{cherti2023reproducible}.
VLMs like CLIP \cite{radford2021learning}, and BASIC \cite{pham2023combined}
focus learning joint text-image representations. 
Dual-encoder models \cite{radford2021learning, pham2023combined} learn context-aware features in a shared latent space \cite{li2020unicoder, doveh2023teaching}, enabling zero-shot image manipulations guided by textual prompts \cite{zheng2022general, YANG2024124183}.
VLMs allow automatic data curation, while SSL potentiates
learning from non-annotated data, forming the crux of our proposal.

\textbf{Our proposed method} 
stands out by autonomously generating high-quality, domain-specific datasets without external supervision or predefined data structures. 
Unlike other synthetic methods \cite{hammoud2024synthclip,tian2024learning},
PaS leverages zero-shot capabilities of LLMs and VLMs to discover domain-specific concepts and collect relevant images. 
This allows us to adapt to any domain without human intervention during the generation process.
PaS focuses on generating domain-specific datasets that are precisely aligned with a target domain, favoring quality over quantity. 
Our datasets are smaller in scale and tailored to capture the nuances of the domain, leading to more efficient training of domain-specific models. 
This is highlighted by our experiments, where models pretrained on PaS datasets outperform those trained on larger, general-domain datasets.
In summary, PaS contributes a novel, scalable, fully autonomous pipeline for domain-specific dataset creation,
addressing the limitations of existing dataset generation methods and highlights the importance of domain-specific data in model pretraining and adaption.\section{Precision at Scale}\label{sec:approach}

\textbf{Precision at Scale (PaS)} is a novel fully automated framework for generating on-demand, domain-specific datasets with minimal human intervention. Our modular pipeline (as shown in \Cref{fig:full-diagram}) comprises three stages:
\textbf{Stage 1} (\Cref{sec:concept_discovery}) utilizes LLMs to discover domain-specific concepts, establishing a foundational concept bank.
\textbf{Stage 2} (\Cref{sec:image_collection}) collects domain-specific images from web-scale databases and generates synthetic images using text-to-image models, enriching the dataset with a broader range of concepts.
\textbf{Stage 3} (\Cref{sec:dataset_curation}) refines the dataset by eliminating image redundancy and filtering irrelevant out-of-domain images through systematic curation techniques.
PaS produces highly precise, scalable,
datasets (\enquote{PaS datasets}) suitable for training self-supervised visual models
or image-text supervision.
Its modular design allows integration with any LLMs, image generators, or image sources, ensuring flexibility across various domains.

\subsection{In-Domain LLM-Guided Concept Discovery} \label{sec:concept_discovery}

\begin{figure*}[!t]
    \centering
    \includegraphics[width=\textwidth]{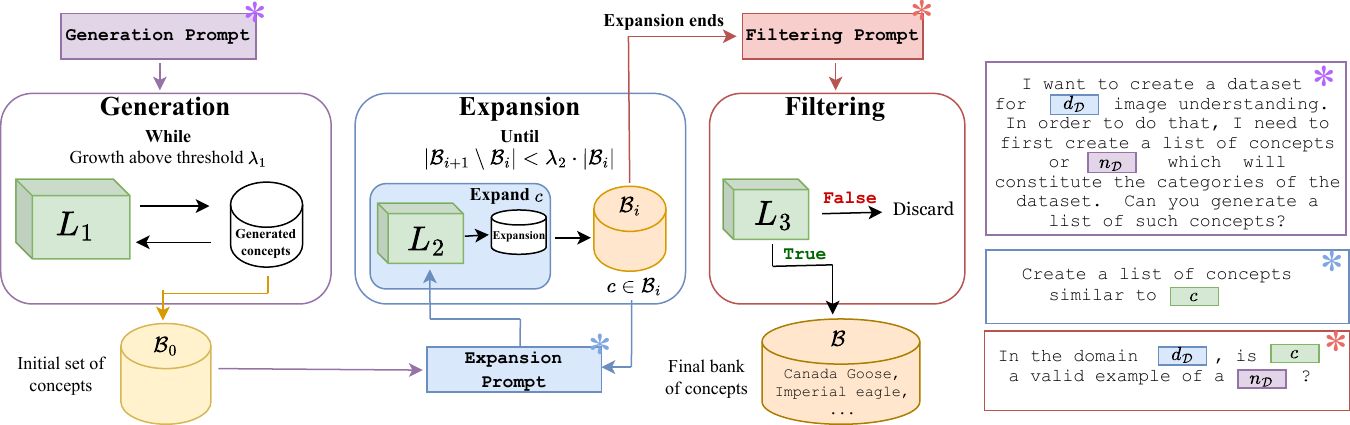}
    \caption{
    \newtextrm{New figure} 
    \textbf{Stage 1
    Workflow:} 
    During the \textbf{generation} step, a prompt template is fed to an LLM, $L_1$, and outputs an initial set of concepts $\mathcal{B}_0$. 
    In the \textbf{expansion} step, a template is used for each concept
    and prompted to the LLM, $L_2$, to find similar concepts.
    Upon maximizing the variety,
    we filter the generated concepts by prompting an auxiliary LLM, $L_3$ (\textbf{filtering} step). 
    resulting in the final bank of concepts $\mathcal{B}$.
    }
    \label{fig:stg1}
\end{figure*}

Stage 1 involves building a comprehensive bank of textual concepts, \(\mathcal{B}\), specific to a target domain, \(\mathcal{D}\). 
The process \textbf{begins with \enquote{only} {\em two inputs}}: the domain name, \(n_{\mathcal{D}}\) (e.g. \enquote{birds}), and a brief description, \(d_{\mathcal{D}}\), of the domain concepts (e.g. \enquote{bird species}). 
This is the only user input required.
This stage (depicted in \Cref{fig:stg1})
consists of three main steps, each
guided by LLMs: 1) generation, 2) expansion, 3) filtering.

\paragraph{Step 1: Generation of Initial Concepts} 
We use an LLM, \(L_1\), to generate an initial set of concepts, \(\mathcal{B}_0\), by prompting it with \(n_{\mathcal{D}}\) and \(d_{\mathcal{D}}\) using the generation prompt template (as shown in \Cref{fig:stg1}). 
To maximize domain coverage, we sample multiple outputs from $L_1$ (with different random seeds) and aggregate them, until the addition of new concepts falls below a threshold $\lambda_1$ (empirically set hyperparameter).
For instance, in the domain of birds, the initial concept set $\mathcal{B}_0$ might include species like 
\enquote{Canada Goose}, 
\enquote{Crow}, 
\enquote{Roseate Spoonbill}, 
\enquote{Broad-billed Sandpiper}, 
and \enquote{Imperial Eagle} (among others).

\paragraph{Step 2: Expansion of Concept Bank}
This step involves expanding and enriching the concept bank within the domain, $\mathcal{D}$.
Each concept in $\mathcal{B}_0$ is expanded using another LLM, $L_2$ (can be same as $L_1$), which generates related concepts by using the expansion prompt template (see \Cref{fig:stg1}).
This is done iteratively: for each concept $c$ in the current set, $\mathcal{B}_i$, $L_2$ generates similar concepts which are aggregated to form the next version of the concept bank, $\mathcal{B}_{i+1}$. The process continues until the growth rate of new concepts falls below a threshold $\lambda_2$ (empirical).
For example, when asked to expand the concept \enquote{Imperial Eagle}, a potential answer of $L_2$ might include elements like \enquote{Bald Eagle}, \enquote{Harpy Eagle},
or \enquote{Golden Eagle}.
The final set of concepts at the end of this expansion is denoted as $\mathcal{B}_{*}$.

\paragraph{Step 3: Concept Filtering}
To exclude irrelevant concepts and minimize computational costs, we use a separate LLM, \(L_3\) (\(L_3 \neq L_1, L_2\)), to validate each concept in \(\mathcal{B}_*\), mitigating potential hallucinations \cite{mundler2024selfcontradictory}. Only concepts confirmed to belong to \(\mathcal{D}\) are kept in the final concept bank \(\mathcal{B}\).

\paragraph{Stage Outcome}
\textit{The output of Stage 1 is the concept bank, $\mathcal{B}$, represented as a set of textual concepts relevant in the domain, $\mathcal{D}$.}

\subsection{Collecting Domain-Specific Images} \label{sec:image_collection}

\begin{figure*}[!t]
    \centering
    \includegraphics[width=0.8\linewidth]{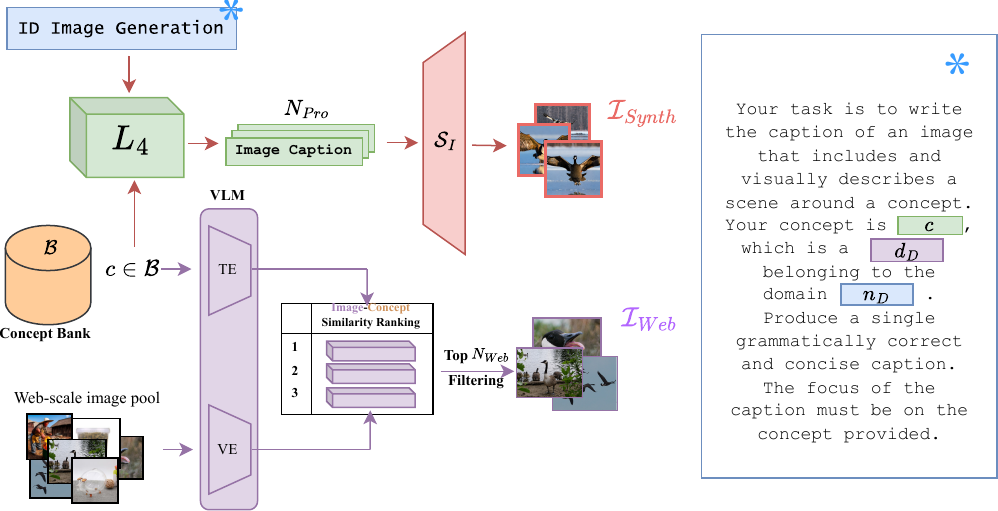}
    \caption{
    \newtextrm{New figure} 
    \textbf{Stage 2 Workflow:} For each valid concept
    from Stage 1, we retrieve $N_{Web}$ most similar images from a web-scale source ($\mathcal{I}_{\text{Web}}$).
    We prompt
    a text-to-image model \( S_I \) using
    LLM-created image captions to produce
    $N_{Pro}\times N_{Synth}$ synthetic images ($\mathcal{I}_{\text{Synth}}$).
    Together they form the unfiltered image dataset, $\mathcal{I}$.
    }
    \label{fig:stg2}
\end{figure*}

Stage 2 (shown in \Cref{fig:stg2}) involves collecting high-quality, domain-specific images using the concept bank $\mathcal{B}$ built in Stage 1.
It has two main components: web-data retrieval and synthetic image generation.

\paragraph{Web-Data Retrieval}
Given a web-scale set of images, $\mathcal{I}_{pool}$, we aim to retrieve the images more aligned with $\mathcal{B}$. For each concept,
we use a VLM to find the images from $\mathcal{I}_{pool}$ that match the concept.
Given $c \in \mathcal{B}$, 
we generate a textual embedding, $\mathbf{t}_c = TE(c)$, using the text encoder $TE(\cdot)$ of a VLM.
We use the visual encoder $VE$ of the VLM to compute the image embedding of each image $I\in\mathcal{I}_{pool}$.
Using cosine similarity, for each concept $c$, we retrieve the $N_{Web}$ most similar images to $t_c$, provided the similarities are larger than a threshold $\lambda_{Web}$  \cite{schuhmann2022laion}.
This results in $\mathcal{I}_{Web}$, a set of web images closely aligned with $\mathcal{B}$.
\newtext{For our experiments, $I_{\text{pool}}$ was sourced from the Re-LAION-5B index. This dataset is a safety-revised version of LAION-5B \cite{schuhmann2022laion} \textit{(updated as of July 2024),} to remove links to any known-illegal content\footnote{\newtext{\url{https://laion.ai/blog/relaion-5b/}}}. Since these images are crawl from the web, they can vary from natural images to sketches. In addition to using this revised dataset, our own retrieval process programmatically respects \texttt{X-Robots-Tag} HTTP header directives, explicitly excluding any images marked with \enquote{noai}, \enquote{noimageai}, \enquote{noindex}, or \enquote{noimageindex} to adhere to content owner preferences.}

\paragraph{In-Domain Synthetic Image Generation}
Instead of using plain concept names for image generation \cite{azizi2023synthetic}, we use an arbitrary LLM, $L_4$, to create detailed prompts that describe hypothetical images for each concept. 
This is achieved using a structured in-domain image generation template (see \Cref{fig:stg2}), which is designed to improve the descriptive quality and diversity of the synthetic images.
For example, for the concept \enquote{Canada Goose}, $L_4$ might generate \enquote{A majestic Canada Goose spreads its wings taking flight above the frozen lake}.
For each concept $c\in\mathcal{B}$, $L_4$ generates $N_{Pro}$ prompts.
These prompts are then used as prompts for a text-to-image model, \( S_I \) (e.g., Stable Diffusion \cite{Rombach_2022_CVPR}), which produces $N_{Synth}$ images per prompt. 
Thus, the total number of synthetic images generated is $|\mathcal{B}|\times N_{Pro} \times N_{Synth}$, forming
$\mathcal{I}_{Synth}$.  
By adjusting $N_{Pro}$ and $N_{Synth}$, we can control the size and diversity of $\mathcal{I}_{Synth}$.

\paragraph{Stage Outcome}
\textit{The final unfiltered image dataset $\mathcal{I}$ is formed by combining web and synthetic images: $\mathcal{I} = \mathcal{I}_{\text{Web}} \cup \mathcal{I}_{\text{Synth}}$.}

\subsection{Autonomous Dataset Curation} \label{sec:dataset_curation}

Stage 3 focuses on refining the unfiltered dataset from
Stage 2, $\mathcal{I}$, to enhance its relevance for $\mathcal{D}$. By systematically eliminating redundant and out-of-distribution (OOD) images, we reduce training costs and minimize potential model performance drawbacks.

\paragraph{Self-Supervised Similarity-based Deduplication}
Duplicate and closely similar images increase image count (and resource consumption) without adding diversity.
We use Self-Supervised Copy Detection (SSCD) \cite{pizzi2022self} to identify and eliminate duplicates.
\torm{This involves computing embeddings for each image using SSCD's visual encoder.
To enhance efficiency, we use the IVF-PQ algorithm \cite{Jegou2011Product} for fast
approximate searches.}%
\newtext{We then build an approximate k-NN graph (k=64) using the IVF-PQ algorithm \cite{Jegou2011Product} for efficient searching\footnote{\newtext{\url{https://rapids.ai/}}}. On this k-NN graph, we then apply a similarity threshold (of 0.6), keeping only the connections between images that exceed a certain cosine similarity. The connected components are then found on this final, pruned graph.}
We retain one (random) representative image per group, resulting in the deduplicated dataset $\mathcal{I}_{\text{dedup}}$.

\begin{figure}[!t]
    \centering
    \subfloat[\newtextrm{New figure} Illustrative 2D Pareto-front example.]{
        \includegraphics[width=0.45\linewidth]{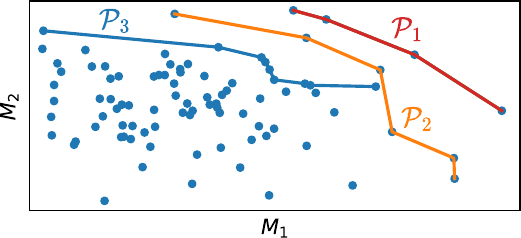}
        \label{fig:pareto-example}
    }\hfill
    \subfloat[\newtext{Illustrative example of $M_3$ computation.}]{
    \includegraphics[width=0.45\linewidth]{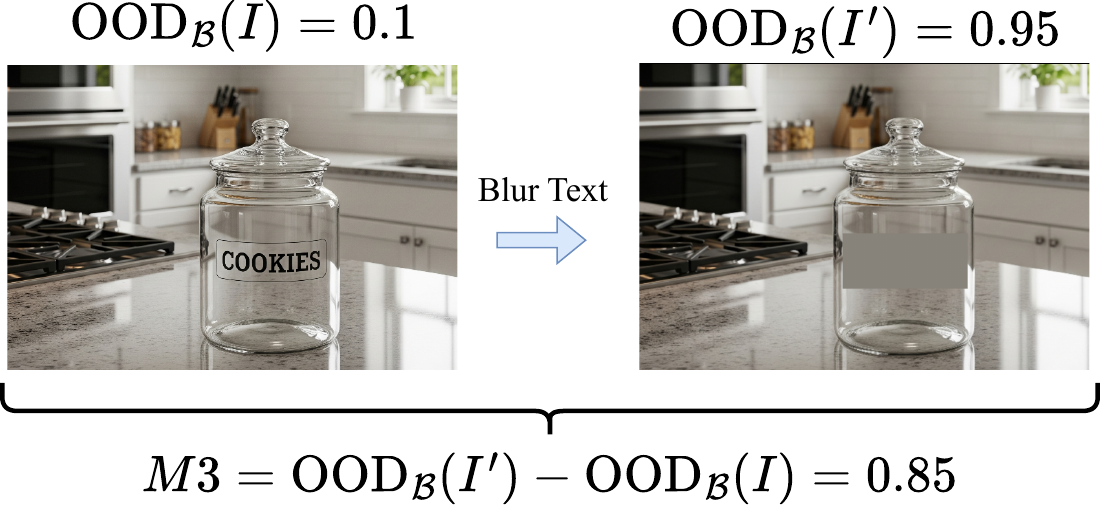}
        \label{fig:m3-example}
    }
    \caption{
    \newtext{Illustrative examples of the Pareto-based filtering.}
    }
    
\end{figure}

\paragraph{CLIP-based OOD Assessment}
Since $\mathcal{I}$ is partially sourced from uncurated sources and generated using unsupervised methods, it \textbf{may contain OOD images.} 
To assess and filter OOD images from
$\mathcal{I}_{dedup}$, we leverage CLIP's zero-shot capabilities, further enhanced by CLIPN \cite{doveh2023teaching}. 
CLIPN utilizes \newtext{learned} dual prompts 
(\textit{positive} prompts to assess the presence and \textit{negative} prompts to assess the absence of a concept)
to
accurately determine the relevance of each image to $\mathcal{D}$. 
\newtext{Specifically, we compute the similarity of each image $I$ with the positive and negative prompt of the concept $c$. The probabilities $p_{c,I},p_{c,I}^{no}$ are then derived by applying softmax to those two similarity scores.}
\textit{We define a 3-dimensional characterization of how much OOD an image is using $M_1$, $M_2$, and $M_3$:}
\begin{itemize}[wide = 0pt]
    \item \textbf{M1: $\text{OOD}_{\mathcal{B}}(I)$}
    calculates the OOD score of image $I$ with respect to specific concept bank $\mathcal{B}$ and is computed as:
    $
    \text{OOD}_{\mathcal{B}} (I) = 1 - \sum_{c \in \mathcal{B}} (1 - p_{c,I}^{no}) \cdot p_{c,I}
    $,
    where $p_{c,I}$ and $p_{c,I}^{no}$ are the probabilities that image $I$ contains and does not contain the concept $c$ respectively.
    
    \item \textbf{M2: $\text{OOD}_{\mathcal{B}'}(I)$}
    assesses the OOD score of image $I$ with respect to general domain descriptors $\mathcal{B}' = \{n_{\mathcal{D}}, d_{\mathcal{D}}\}$.
    $
    \text{OOD}_{\mathcal{B}'} (I) = 1 - \sum_{c \in \mathcal{B}'} (1 - p_{c,I}^{no}) \cdot p_{c,I}
    $.

    \item \textbf{M3: Text Influence Mitigation}
    helps in reducing biases introduced by textual elements that may falsely align an image with the target domain based solely on text presence \cite{maini2024tmars}.
    To mitigate this influence,
    we define $M_3$:
    $
    \text{OOD}_{\mathcal{B}}(I') - \text{OOD}_{\mathcal{B}}(I)
    $,
    where $I'$ is obtained by
    using a text-detection algorithm to detect and blur text regions in $I$.
    \newtext{A high $M_3$ score, indicating that the image is only relevant due to its text, heavily penalizes it during Pareto-front removal (as exemplified in \Cref{fig:m3-example}).}
\end{itemize}

\paragraph{Pareto Front-Based Removal} 
We employ a Pareto front-based multi-objective optimization approach
which effectively ranks and eliminates OOD images based on the computed OOD metrics $M_1$, $M_2$, and $M_3$. Instead of prioritizing one single factor, we simultaneously consider all relevant factors to identify and rank the least relevant images. 
We show a simplified 2D Pareto fronts illustration in \Cref{fig:pareto-example}, where each point represents an image and the first three fronts ($\mathcal{P}_1$, $\mathcal{P}_2$, $\mathcal{P}_3$) are highlighted. 
The Pareto Front removal process is iterative and includes:
\begin{enumerate}[wide = 0pt, label={(\arabic*)}]
    \item 
    \textbf{Identify Pareto Front}: Determine the first Pareto front \( \mathcal{P}_1 \) within the deduplicated dataset \( \mathcal{I}_{dedup} \). The Pareto front consists of images that are the most OOD across all metrics. An image is part of the Pareto front if there is no other image that simultaneously performs \textit{worse} in all $M_1, M_2, M_3$. In other words, these images are among the worst in at least one metric and not better in any other metric.
    \item 
    \textbf{Remove Pareto Front}: Exclude \( \mathcal{P}_1 \) set from \( \mathcal{I}_{dedup} \).
    \item 
    \textbf{Repeat}: Recompute next Pareto front \( \mathcal{P}_2 \) on the updated dataset and remove it. Continue this process iteratively (identifying \& removing \( \mathcal{P}_3 \), \( \mathcal{P}_4 \), etc.) until the dataset meets desired size or quality threshold.
\end{enumerate}

\paragraph{Stage Outcome}
\textit{Pareto-Front ensures that the most irrelevant OOD images are excluded first, maintaining the relevance and quality of the dataset, \textit{leading to the final dataset, $\mathcal{I}_{final}$.}}

\subsection{Final Curated Dataset}

The final output of PaS
is a meticulously curated, domain-specific dataset, assembled autonomously without human oversight. 
PaS datasets are ideal for self-supervised pretraining of visual models, as well as fine-tuning existing VLMs, thanks to the image-text pairs obtained naturally from the PaS pipeline. 
These high-quality datasets enable the development of specialized models that excel in their respective domains.
In the following sections, we assess the richness and suitability of
PaS datasets (\Cref{sec:results-task-agnostic}) and evaluate their effectiveness as SSL pretrainers and VLM \enquote{fine-tuners} (\Cref{sec:results-downstream}).
\section{PaS Dataset Diversity \& Domain Assessment}\label{sec:results-task-agnostic}

The primary goal of PaS is to create diverse and useful domain-specific datasets.
Beyond curation, we include a low-resource task-agnostic technique for assessing the dataset diversity.
In this section, we detail each component by evaluating our PaS
datasets
against well-established domain-specific supervised SoTA datasets - Food-2K \cite{min2023large}, iNat\textsubscript{Birds} \cite{van2018inaturalist}, AMI-GBIF \cite{jain2024insect} for \textit{food}, \textit{birds}, and \textit{insects}, respectively.
For each domain, we create a corresponding PaS dataset. (PaS-F for \textit{food}, PaS-B for \textit{birds}, PaS-I for \textit{insects}).

\subsection{Dataset Overview.}
 
\begin{figure*}[!t]
    \centering
    \includegraphics[width=0.99\linewidth]{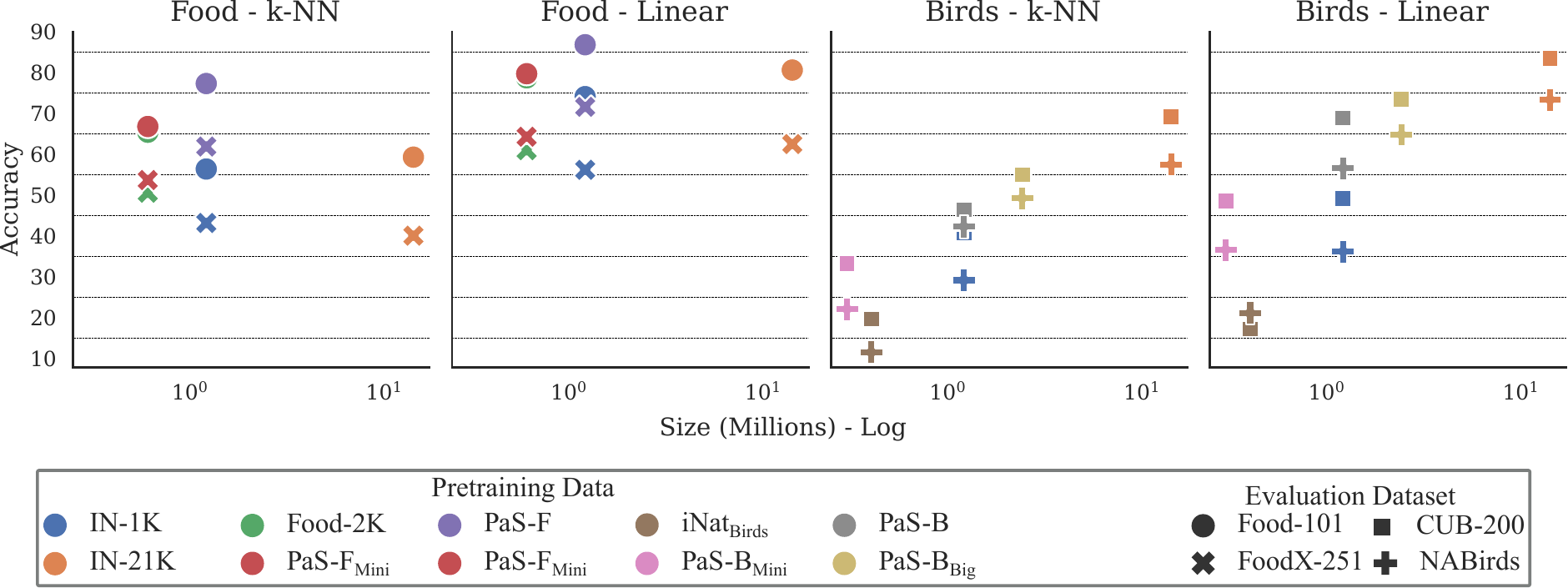}
    \caption{%
    \textbf{\torm{PaS size and kNN performance}}\newtext{\textbf{K-NN Top-1 classification accuracy} (in \%) of different pretraining datasets against the pretraining dataset size (in log scale)}.}
    \label{fig:pas-evolution}
\end{figure*}

PaS datasets, in general, can be created for any given size.
For effective comparisons, we use IN-1K and SoTA domain datasets as size references for our datasets.
Only for \textit{food} and \textit{birds}, we create mini-versions, PaS-F\textsubscript{Mini} and PaS-B\textsubscript{Mini} as both Food-2K and iNat\textsubscript{Birds} are much smaller than IN-1K.
AMI-GBIF on the other hand is larger
to IN-1K (PaS-I$\approx$1.9M images).
As shown in \Cref{tab:pas-evolution}, the number of concepts in PaS datasets often exceeds that (classes) of domain-specific datasets (Food-2K has 2K, iNat\textsubscript{Birds} has 1,486, AMI-GBIF has $>$5K classes), highlighting the diversity of our concept bank. 
As visualized in \Cref{fig:pas-evolution}, PaS datasets show stronger performances as pretrainers on the same scale as domain-specific supervised datasets and also being way smaller than general-purpose datasets like IN-21K (one order smaller), highlighting the effectiveness of our curation pipeline. Detailed results can be found in \Cref{sec:results-downstream}.

\begin{table}[t!]
    \centering
    \renewcommand{\arraystretch}{0.8}
    \resizebox{\linewidth}{!}{%
    \begin{tabular}{l|cccc}
        \toprule
        & \textbf{Concepts} & \textbf{Raw Images} & \textbf{Final Images} & \textbf{Synth/Real\%} \\
        \midrule
        \textcolor{color1}{PaS-F} & \textcolor{color1}{3.3k} & \textcolor{color1}{1.6M} & \textcolor{color1}{1,177k} & \textcolor{color1}{60/40} \\
        \textcolor{color2}{PaS-B} & \textcolor{color2}{5.0k} & \textcolor{color2}{1.5M} & \textcolor{color2}{1,200k} & \textcolor{color2}{46/54} \\
        \textcolor{color3}{PaS-I} & \textcolor{color3}{16.6k} & \textcolor{color3}{2.5M} & \textcolor{color3}{1,904k} & \textcolor{color3}{90/10} \\
        \textcolor{color4}{PaS-F\textsubscript{Mini}} &  \textcolor{color4}{3.3k} & \textcolor{color4}{1.6M} & \textcolor{color4}{627k} & \textcolor{color4}{86/14} \\
        \textcolor{color5}{PaS-B\textsubscript{Mini}} &  \textcolor{color5}{5.0k} & \textcolor{color5}{1.5M} &\textcolor{color5}{450k} & \textcolor{color5}{86/14} \\
        \textcolor{color6}{PaS-B\textsubscript{Big}} &  \textcolor{color6}{5.0k} & \textcolor{color6}{3.2M} &\textcolor{color6}{2,421K} & \textcolor{color6}{49/51} \\
        \bottomrule
    \end{tabular}%
    }
    \caption{Statistics of the different PaS datasets evaluated in this paper. \textit{Colors represent markers in \Cref{fig:pas-evolution}}.}\label{tab:pas-evolution}
\end{table}

\subsection{Diversity and Domain Alignment.}
We assess the suitability of PaS datasets
by addressing the question:
\textbf{Does a PaS dataset effectively represent the target domain?}
To answer this, we propose a framework with three distinct perspectives:
(1) distribution of lexical concepts, 
(2) distribution of image embeddings, 
(3) semantic richness of concept-level prototypes.
We evaluate our autonomous PaS
datasets under each perspective without any specific task.
For each assessment, we compare the PaS datasets with SoTA-supervised datasets (Food-2K, iNat\textsubscript{Birds}, AMI-GBIF) and other popular domain-specific benchmarks in each domain (Food-101 \cite{bossard2014food} and FoodX-251 \cite{kaur2019foodx} for \textit{food}, CUB-200-2011 \cite{wah2011caltech} and NABirds \cite{Horn_2015_CVPR} for \textit{birds}, and different official splits of AMI-GBIF for \textit{insects}).

\begin{figure*}[!htpb]
    \centering
    \subfloat[Lexical space.]{
        \includegraphics[width=0.99\textwidth]{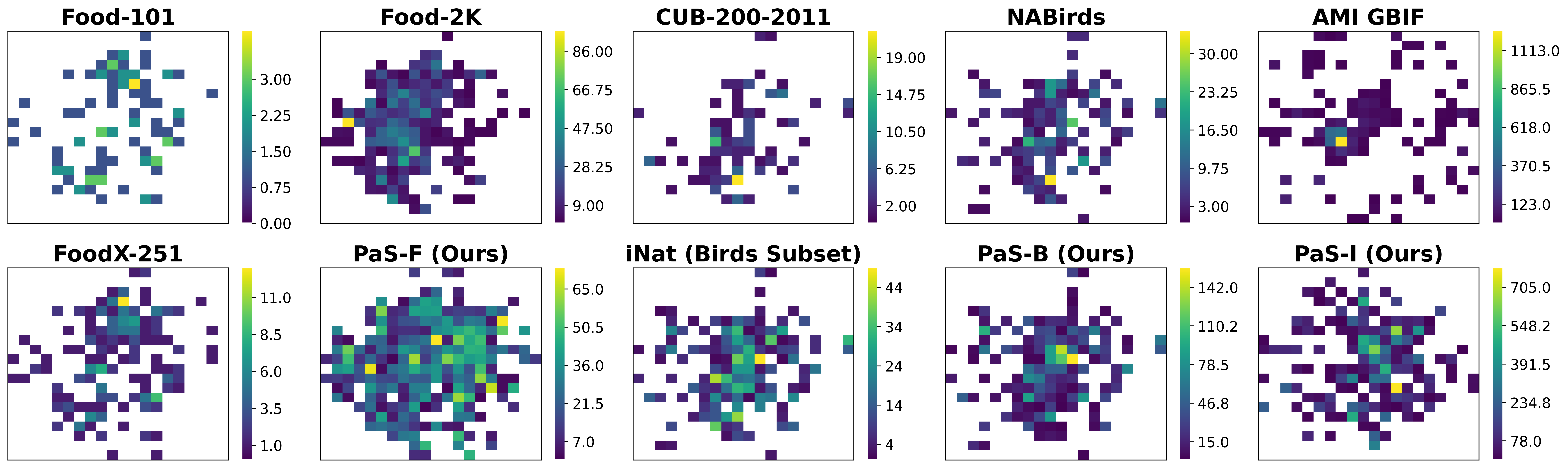}
        \label{fig:full-galaxies-all-lexical}
    }\\
    \subfloat[\newtext{Lexical space, comparison on the same scale. In each column, the number of concepts in the PaS dataset has been randomly subsampled to match the size of the vocabulary of the supervised dataset.}]{
        \includegraphics[width=0.99\textwidth]{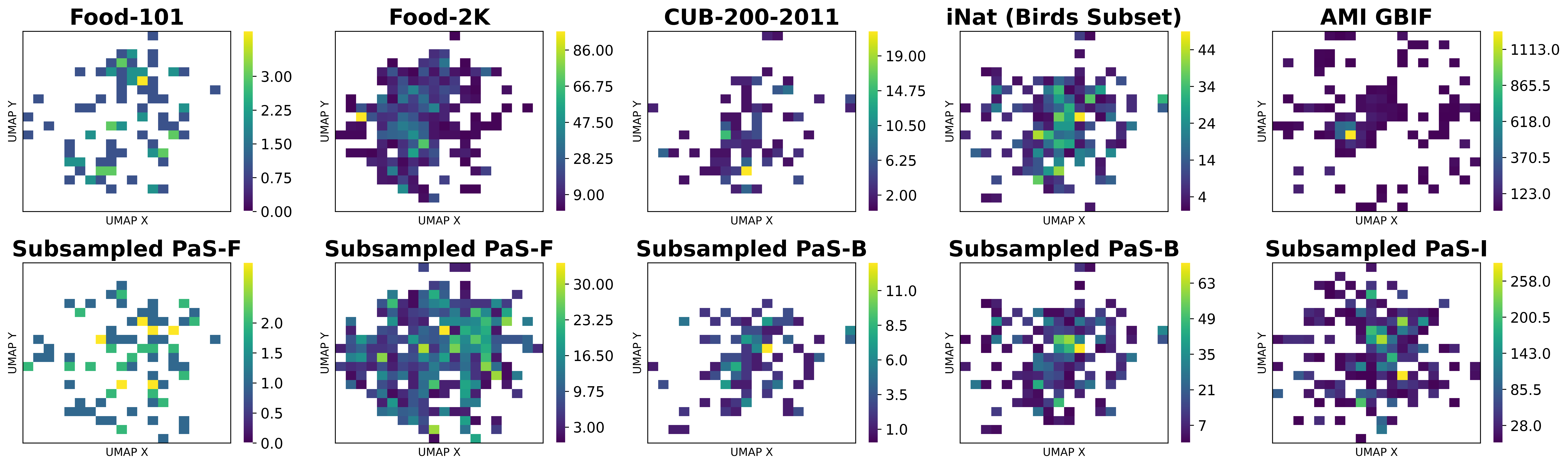}
        \label{fig:subsampled-galaxies-all-lexical}
    }\\
    \subfloat[Image embeddings.]{
        \includegraphics[width=0.99\textwidth]{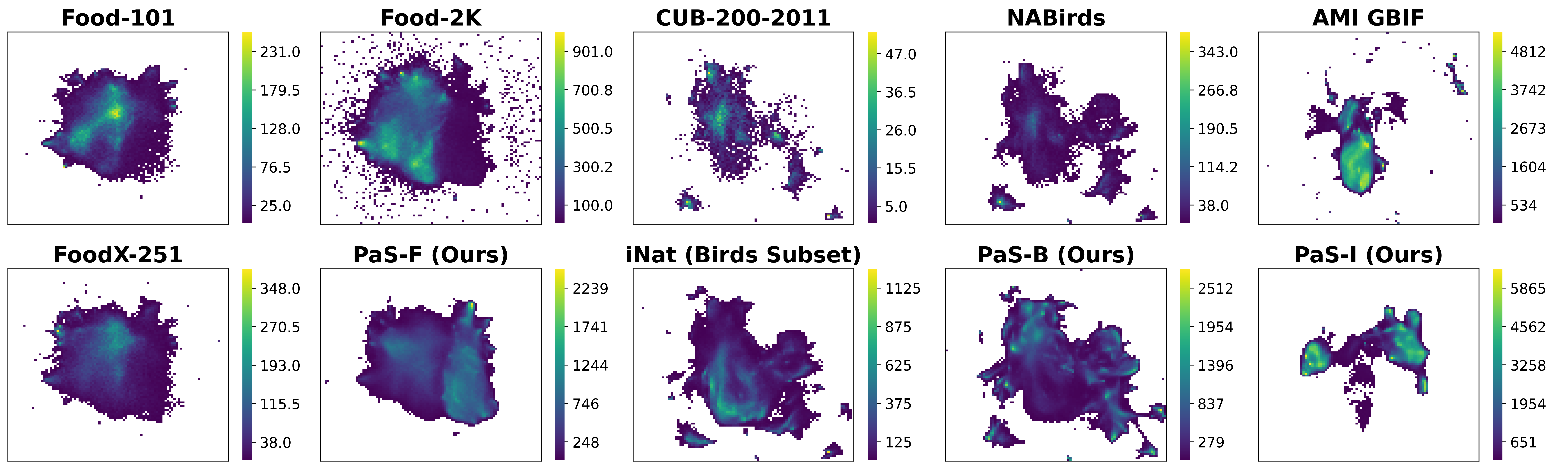}
        \label{fig:full-galaxies-all-img}
    }\\
    \subfloat[\newtext{Visual space, comparison on the same scale. In each column, the number of images in the PaS dataset has been randomly subsampled to match the size of the supervised dataset.}]{
        \includegraphics[width=0.99\textwidth]{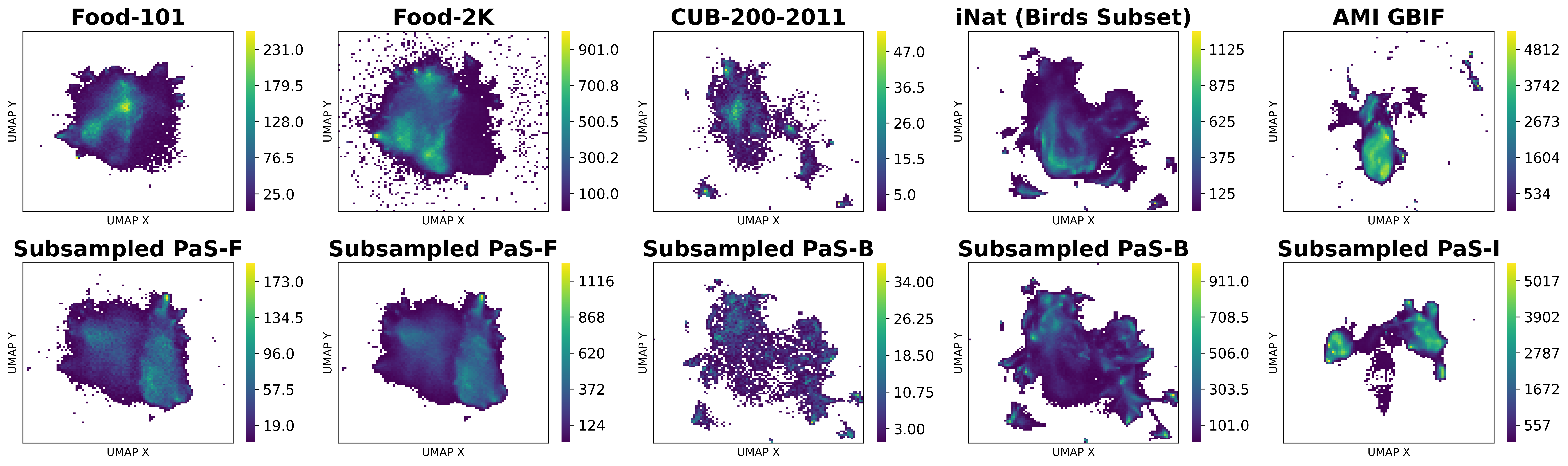}
        \label{fig:subsampled-galaxies-all-visual}
    }
    \caption{Density maps of the UMAP representations for the lexical space of concepts and image embeddings across all domain-specific datasets. \newtext{The scale indicates concept or image concentration in the latent space.}}
    \label{fig:omega_galaxy}
\end{figure*}

\subsubsection{Distribution of Lexical Concepts.}
LLM-generated concept bank determines the diversity of PaS datasets and should exclusively cover the target domain.
To highlight the coverage of domains, we compute lexical embeddings of concepts  and labels using OpenAI's CLIP ViT-L/14 text encoder \cite{radford2021learning}
and compare the distributions using a reduced UMAP space \cite{mcinnes2018umap}.
The process involves the following steps:
\begin{enumerate}[wide = 0pt, label={(\arabic*)}, topsep=2pt, itemsep=1pt, parsep=0pt]
    \item Encoding dataset concepts into latent representations using CLIP.
    \item Reducing the dimensionality of these embeddings to 2D using UMAP.
    \item Dividing the resulting 2D UMAP space into a uniform grid.
    \item Creating density maps for each dataset by counting the number of concepts within each grid cell.
\end{enumerate}
We show density maps of lexical concepts for all domains in \Cref{fig:full-galaxies-all-lexical}.
We opted for density maps over scatter plots to prevent densely populated regions from being obscured, thereby providing a clearer visualization of the concept distribution across different datasets.
\newtext{To prove that the enhanced coverage of PaS datasets is not merely an effect of size, we obtained the same plots after subsampling our concepts to match the scale of the supervised datasets (\Cref{fig:subsampled-galaxies-all-lexical}).}
In \textit{food}, the lexical distribution demonstrates that the concepts in PaS-F extensively cover the embedding space,
effectively bridging the gaps between classes from different datasets.
In \textit{birds}, concepts of PaS-B exhibits a wide and varied distribution across the embedding space compared to CUB-200-2011 and NABirds, and it
closely matches the granularity of iNat\textsubscript{Birds}.  
This high overlap
highlights that the generated concepts align well with the target domain, indicating an effective concept generation.
In \textit{insects}, 
we see less overlap. 
The fact that AMI-GBIF covers mostly moth species, a subset of all insects (target of PaS-I) could be a possible reason. 
Overall, the broad \textbf{coverage and significant alignment} across domains \newtext{in both \Cref{fig:full-galaxies-all-lexical,fig:subsampled-galaxies-all-lexical}} underscore the ability of PaS-generated concepts to enrich dataset diversity and relevance to specific domains.

\subsubsection{Distribution of Image Embeddings.}
Similar to the analysis of lexical concepts, we also compare the image distributions of the different datasets. This process is completely analogous to the previous one. The main differences are that, in this case, we obtain an embedding per image (not per class) and that we use a ResNet-152 pretrained on IN-21K to compute embeddings. 
We then compare the latent distributions for each domain, which can be seen in \Cref{fig:full-galaxies-all-img}.

\begin{figure}[!htpb]
    \centering
    \includegraphics[width=\linewidth]{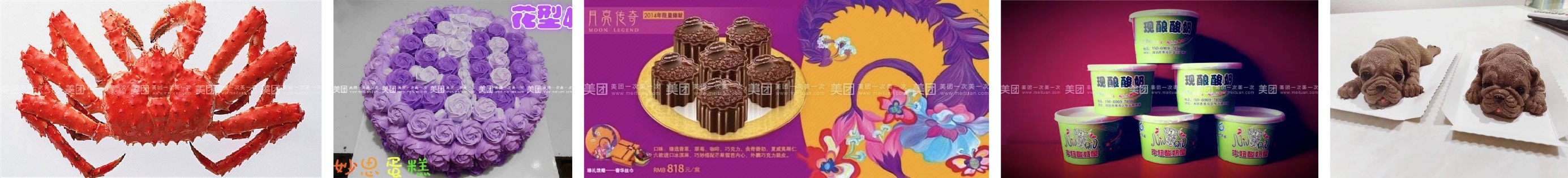}
    \caption{\newtext{Examples of OOD images in Food-2K dataset, identified in \Cref{fig:full-galaxies-all-img}.}}
    \label{fig:ood_food2k}
\end{figure}

A good alignment and span in feature space
would prove a comprehensive coverage of the target domain.
In both \textit{food} and \textit{birds}, there is a notable alignment across all datasets. The impact of larger datasets, which exhibit fewer gaps in the visual space, is particularly evident in PaS-B. PaS-B achieves the most extensive coverage of the embedding space among the considered datasets.
Upon analyzing the density distribution, we observe that while the CUB-200-2011 dataset has densely populated regions, PaS-B,
along with others, display a more uniform distribution across the embedding space. This uniformity is attributed to two key factors:
    (1) A reduced number of images in CUB-200-2011, and,
    (2) A limited variety of bird species in CUB-200-2011 compared to other datasets.
Similarly, in \textit{food}, Food-2K spans a broader area but includes numerous outliers, potentially indicating OOD images \newtext{(see \Cref{fig:ood_food2k} for some outliers examples)}. 
In contrast, PaS-F encompasses the embedding spaces of both Food-101 and Food-2K. In particular, PaS-F, exhibits a uniform distribution of embeddings that is well-balanced between densely covered and sparsely populated regions, which
suggests a comprehensive representation of the \textit{food} domain.
Regarding \textit{insects}, while the overall shape of the distribution aligns across datasets (PaS-I being slightly broader), the densities vary significantly.
Sample-wise, under-representation of certain images and semantics in AMI-GBIF might cause potential differences.
Specifically, most images in AMI-GBIF
are concentrated in a particular area of the embedding space. A closer examination of the semantics of each dataset provides further insights into this behavior (see \Cref{sec:semantic-richness-analysis}).
Overall, these observations underscore the effectiveness of PaS in automatically creating large-scale, domain-specific datasets in terms of image coverage and alignment with existing datasets. \newtext{This is not just an artifact of their larger size, as the trend holds when our datasets are subsampled to an equal number of images (\Cref{fig:subsampled-galaxies-all-visual}), highlighting a more diverse collection of visual features.}

\subsubsection{Semantic Richness Analysis}
\label{sec:semantic-richness-analysis} 

Semantic richness refers to the depth and diversity of concepts and information captured within a dataset, and evaluates 
how well a dataset captures the nuances, variations, and comprehensive coverage of concepts within a specific domain. 
A \enquote{prototype} \cite{van2023prototype} is a
representative semantic 
that captures the most salient features of a concept, making them useful for understanding data diversity and exhaustiveness.

\begin{figure*}[tp]
    \includegraphics[width=\linewidth]{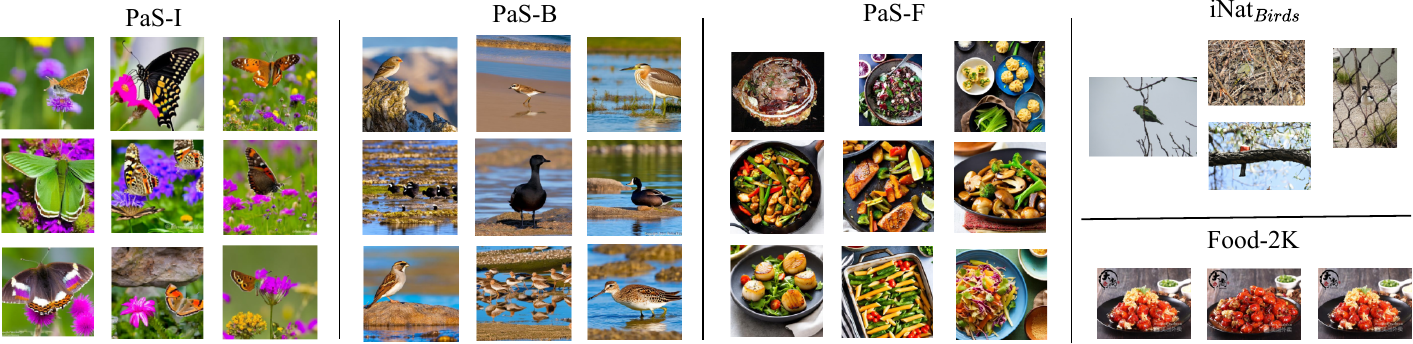}
    \caption{
    \textbf{Extracted exclusive prototypes examples.} 
    Each image belongs to an exclusive prototype.
    }\label{prototypes}
\end{figure*}

\begin{table}[]
    \centering
    \renewcommand{\arraystretch}{0.8}
    \begin{tabular}{lcc}
    \hline
    Dataset    & Exclusive & Images \\ \hline
    Food-2K     & 16        & 121    \\
    PaS-F      & \textbf{32}        & \textbf{310}    \\\hline
    iNat\textsubscript{Birds} & 10        & 25     \\
    PaS-B      & \textbf{23}        & \textbf{358}    \\ \hline
    AMI        & 0         & 0      \\
    PaS-I      & \textbf{4}         & \textbf{650}    \\ \hline
    \end{tabular}
    \captionof{table}{Exclusive prototypes for PaS and SoTA supervised datasets.}
    \label{tab:prototypes}
\end{table}

We show examples of the most relevant exclusive prototypes for each dataset in \Cref{prototypes} and the number of unique prototypes with image count in \Cref{tab:prototypes}.
\textit{Birds} has 8159 shared prototypes,
indicating a substantial overlap between datasets. 
Comparing unique prototypes,
PaS-B contains much greater diversity both in the prototypes and the population of those prototypes.
\textit{Food} has 8144 shared prototypes.
Food-2K is human-labelled but still contains several repetitions that fill its exclusive prototypes.
Exclusive prototypes of PaS-F
highlights the diversity not included in Food-2K.
In \textit{insects}, there are 8188 shared prototypes.
PaS-I contains all the semantics available in AMI-GBIF, evidenced by 
not identifying any AMI-exclusive prototype.
While the number of samples per shared prototype might vary, we can confidently claim that PaS-I manages to match and improve the diversity provided by AMI-GBIF. Comparing the most significant shared prototype, we find that the most populated AMI-GBIF prototype represents a human-driven concept (quantity over diversity). \textcolor{black}{The differences 
in the behaviour of distributions found in previous sections can be explained by the prevalence of this prototype in the AMI dataset. 
More details are provided in 
\ref{sec:appendix-insects}.
}

\subsection{Illustrative Examples of PaS Dataset Curation}

\begin{figure*}[tp]
    \centering
    \hspace*{\fill}%
    \subfloat[PaS-F]{{\includegraphics[height=6.5cm]{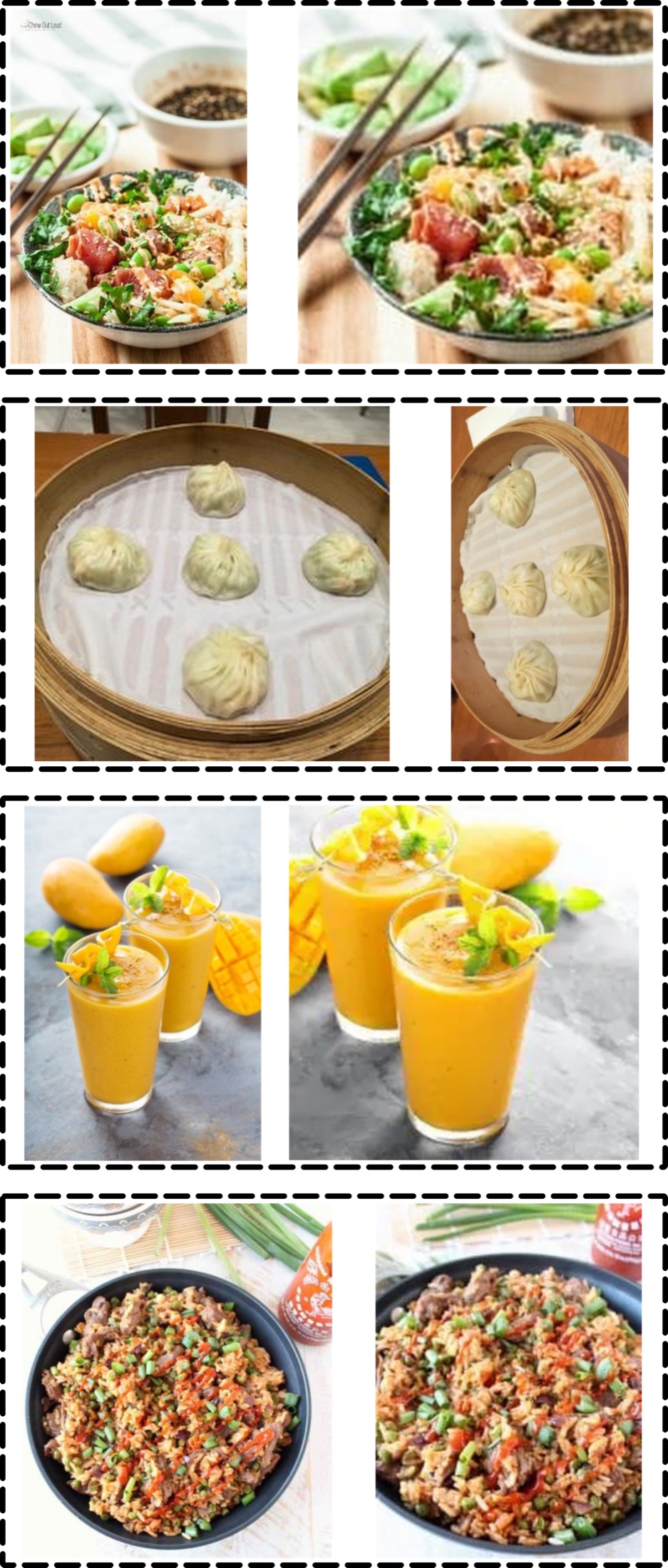}}\label{fig:deduplication_food}}%
    \hspace{1.5em}
    \subfloat[PaS-B]{{\includegraphics[height=6.5cm]{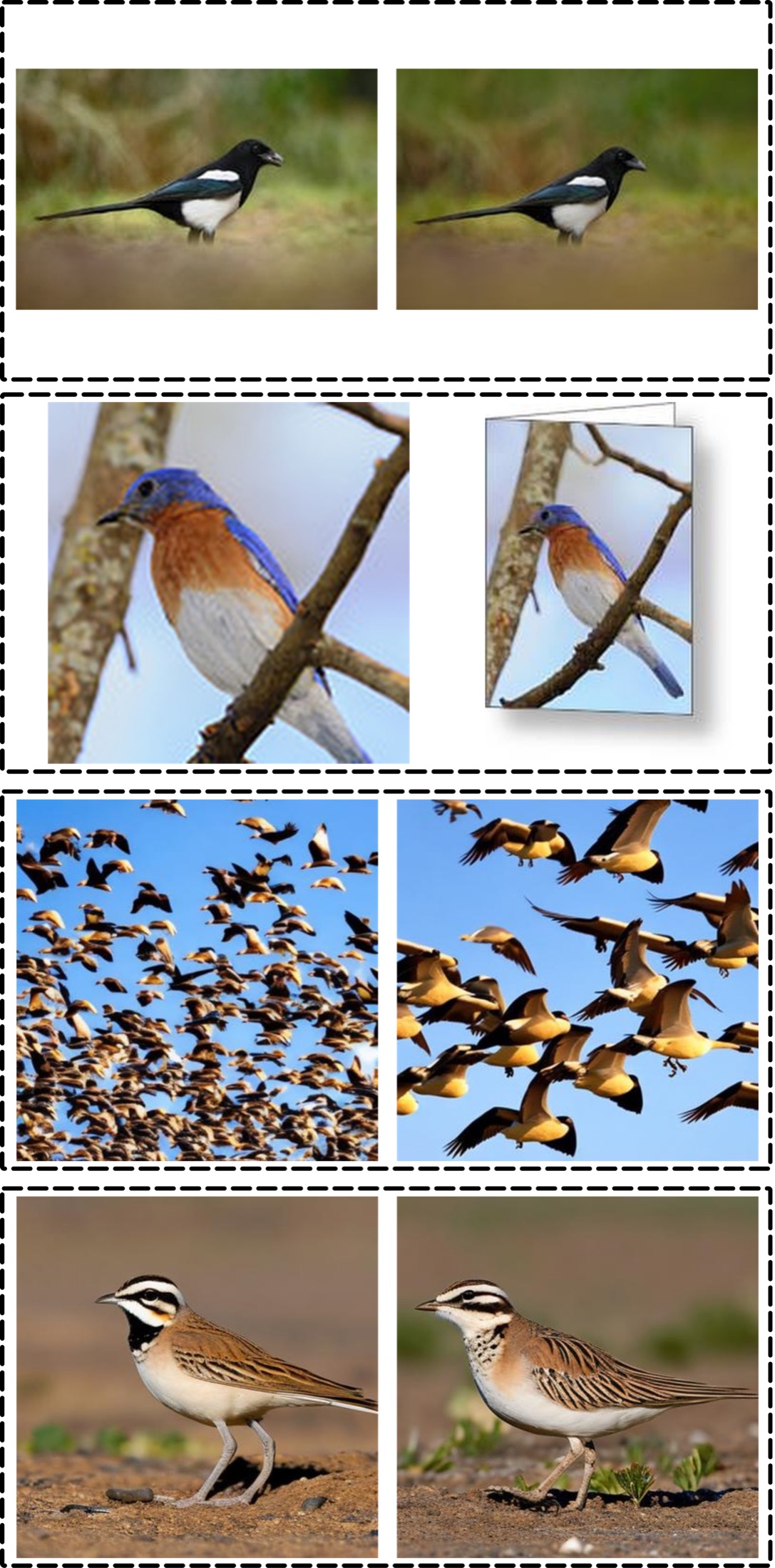}}\label{fig:deduplication_birds}}%
    \hspace{1.5em}
    \subfloat[PaS-I]{{\includegraphics[height=6.5cm]{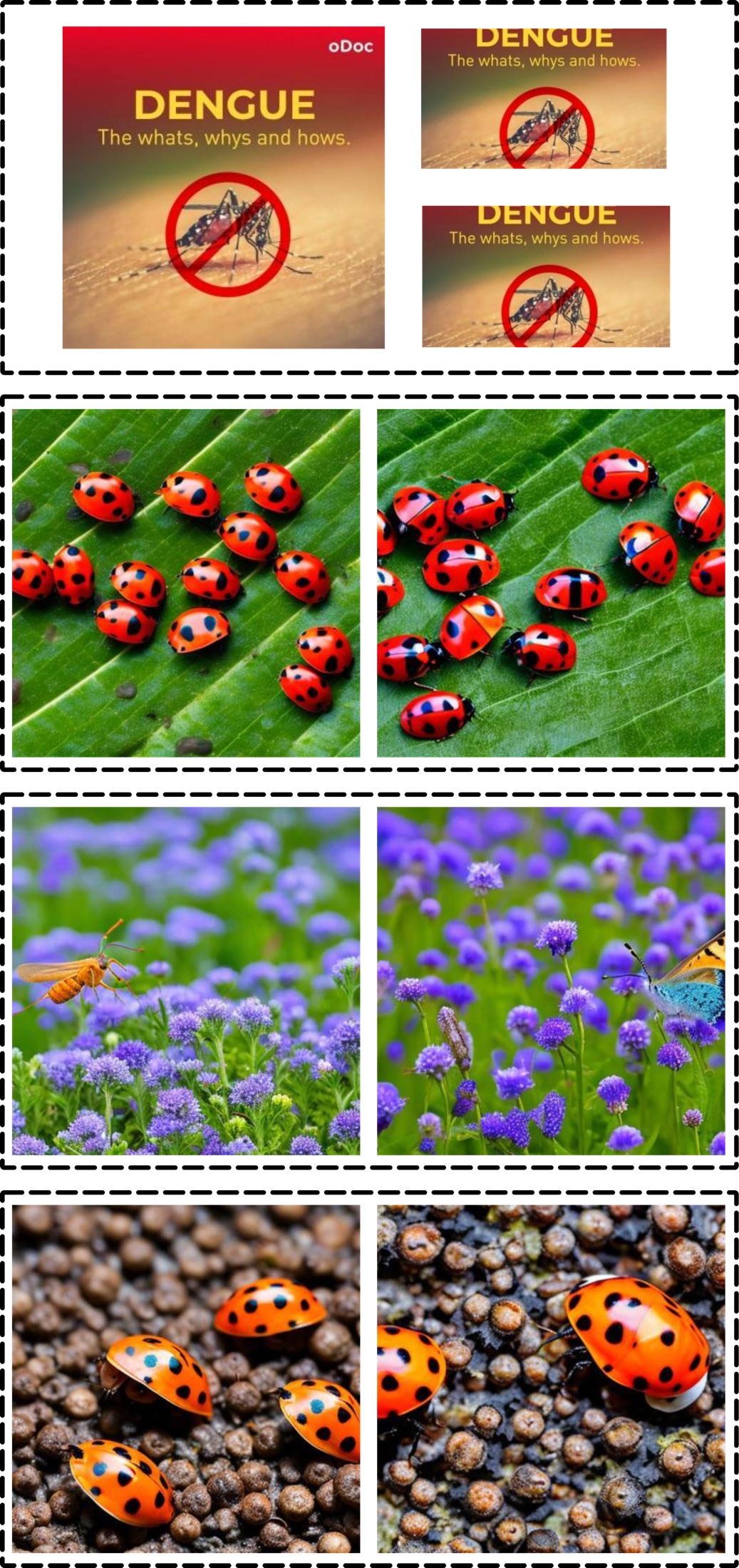}}\label{fig:deduplication_insects}}%
    \hspace*{\fill}

    \caption{Examples of duplicates found and removed in the creation of the PaS datasets.}
    \label{fig:deduplication_examples}
\end{figure*}

\begin{figure*}[tp]
    \centering
    \hspace*{\fill}%
    \subfloat[PaS-F]{{\includegraphics[height=6cm]{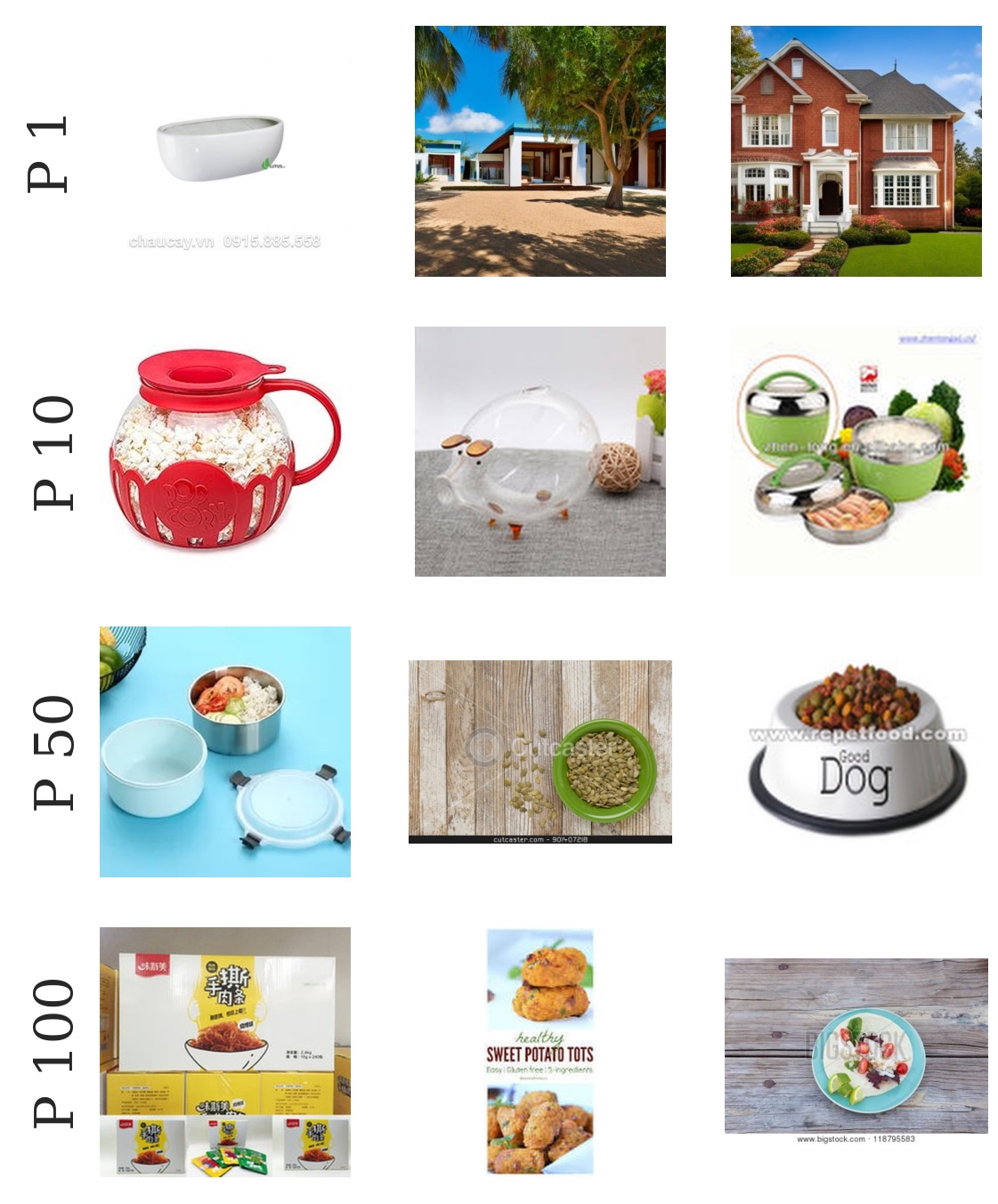}}\label{fig:pareto_food}}%
    \hspace{1.5em}
    \subfloat[PaS-B]{{\includegraphics[height=6cm]{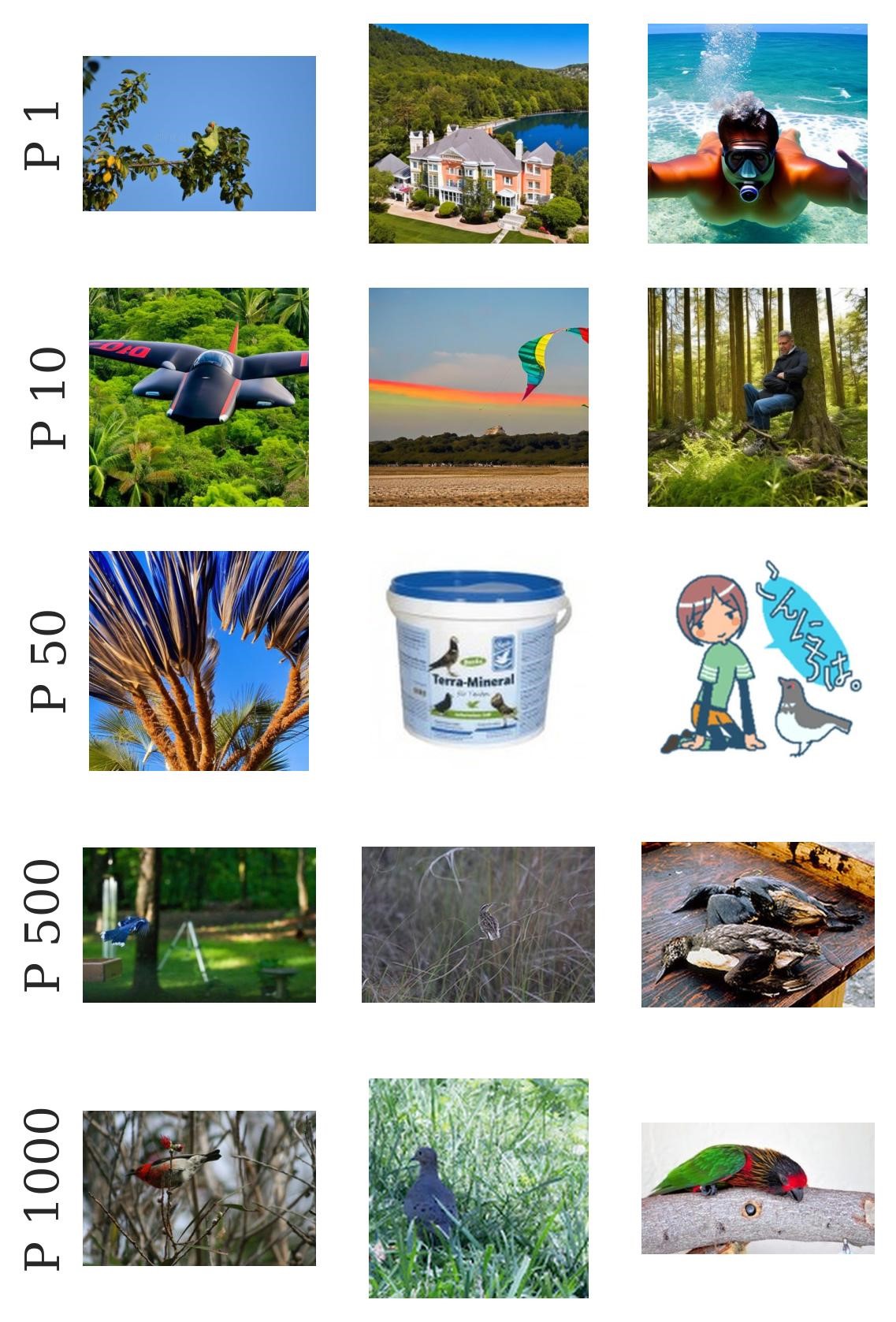}}\label{fig:pareto_birds}}%
    \hspace{1.5em}
    \subfloat[PaS-I]{{\includegraphics[height=6cm]{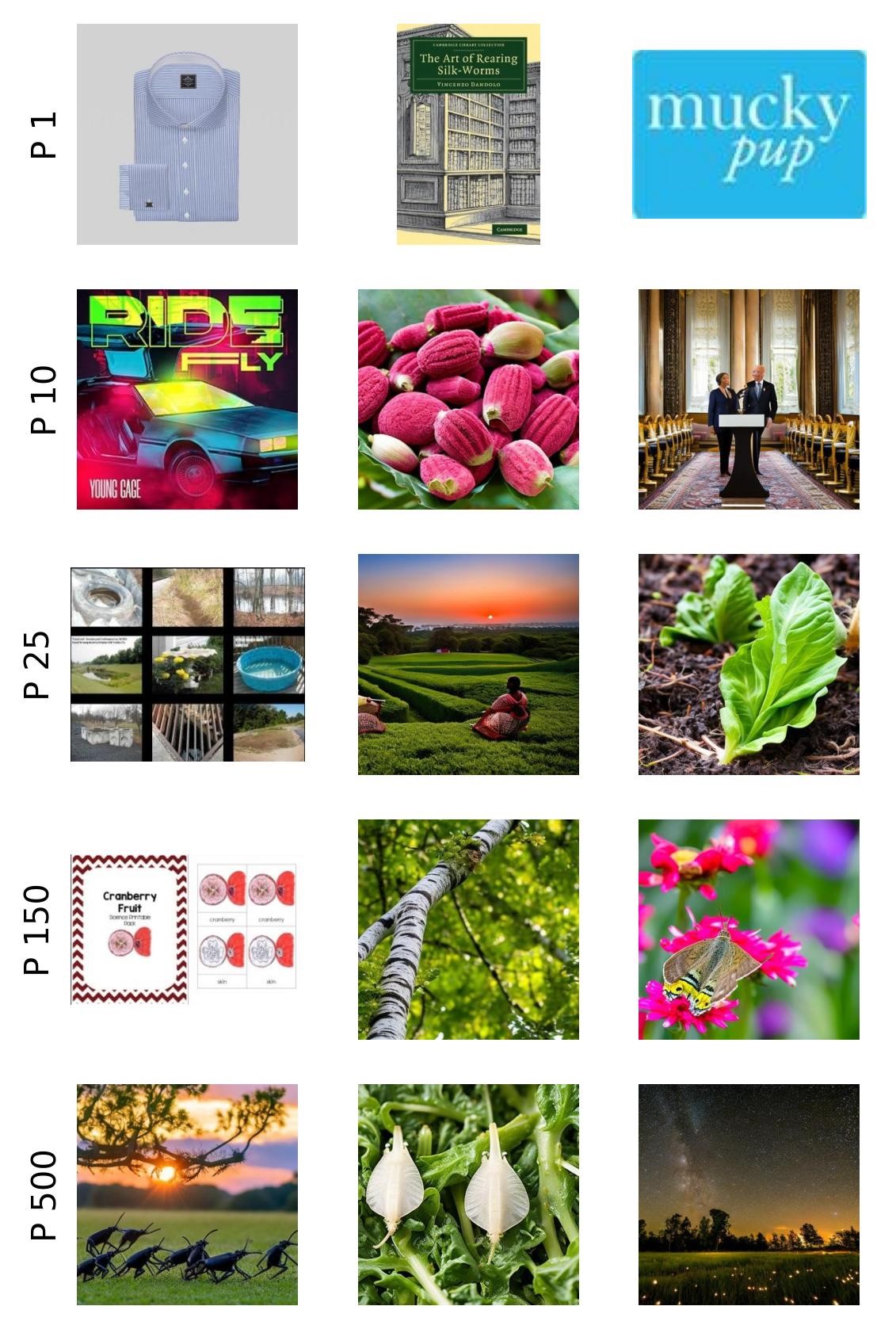}}\label{fig:pareto_insects}}%
    \hspace*{\fill}
    \caption{\textbf{Examples of images removed during the Pareto-based removal.} \enquote{P \textit{n}} refers to images belonging to the $n$-th Pareto front (removed in the $n$-th iteration).}
    \label{fig:pareto_examples}
\end{figure*}

\begin{figure*}[tp]
    \centering
    \makebox[\linewidth]{%
    \hfill
    \subfloat[PaS-F]{{\includegraphics[height=8.28cm]{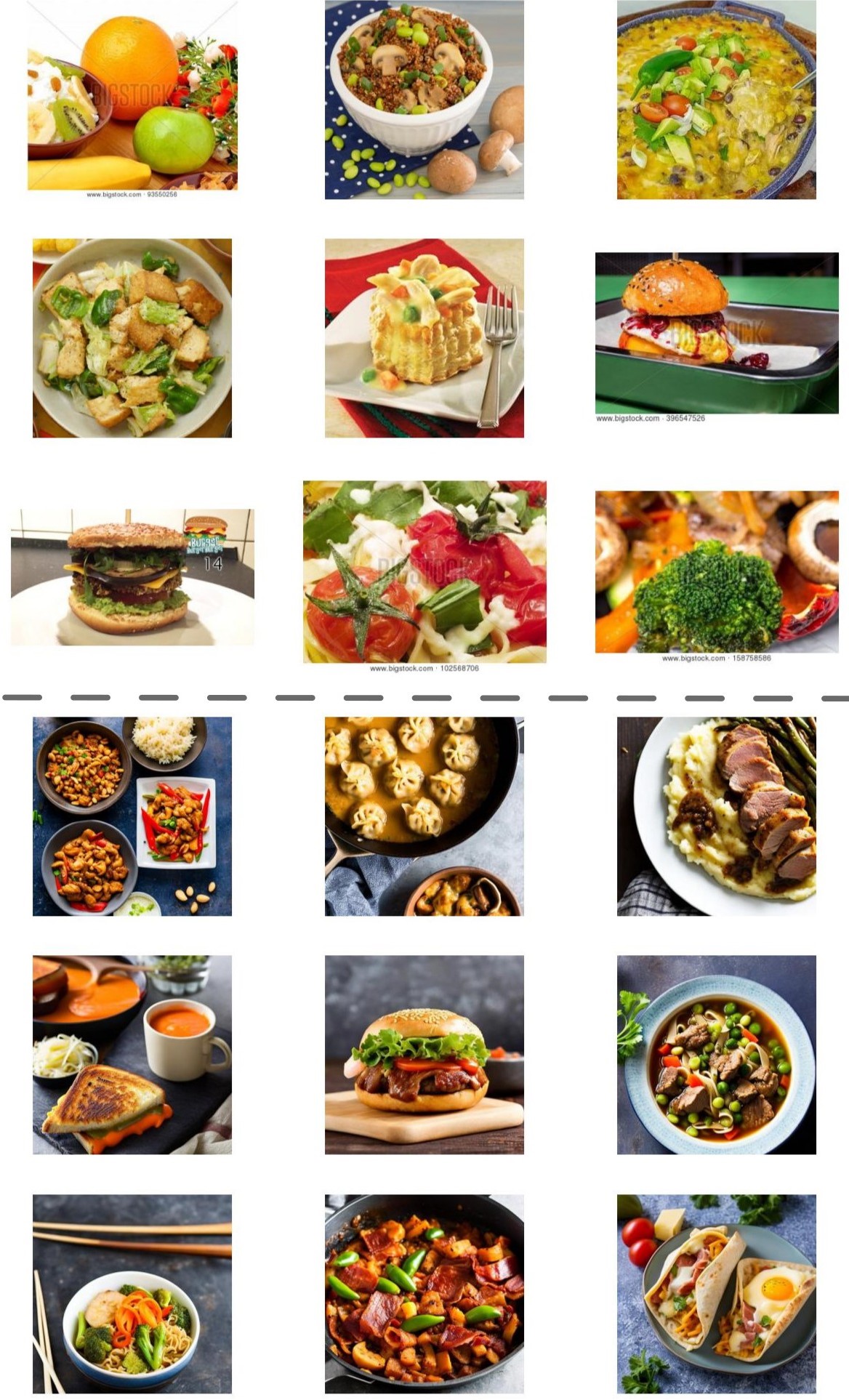}}}%
    \hfill
    \subfloat[PaS-B]{{\includegraphics[height=8.28cm]{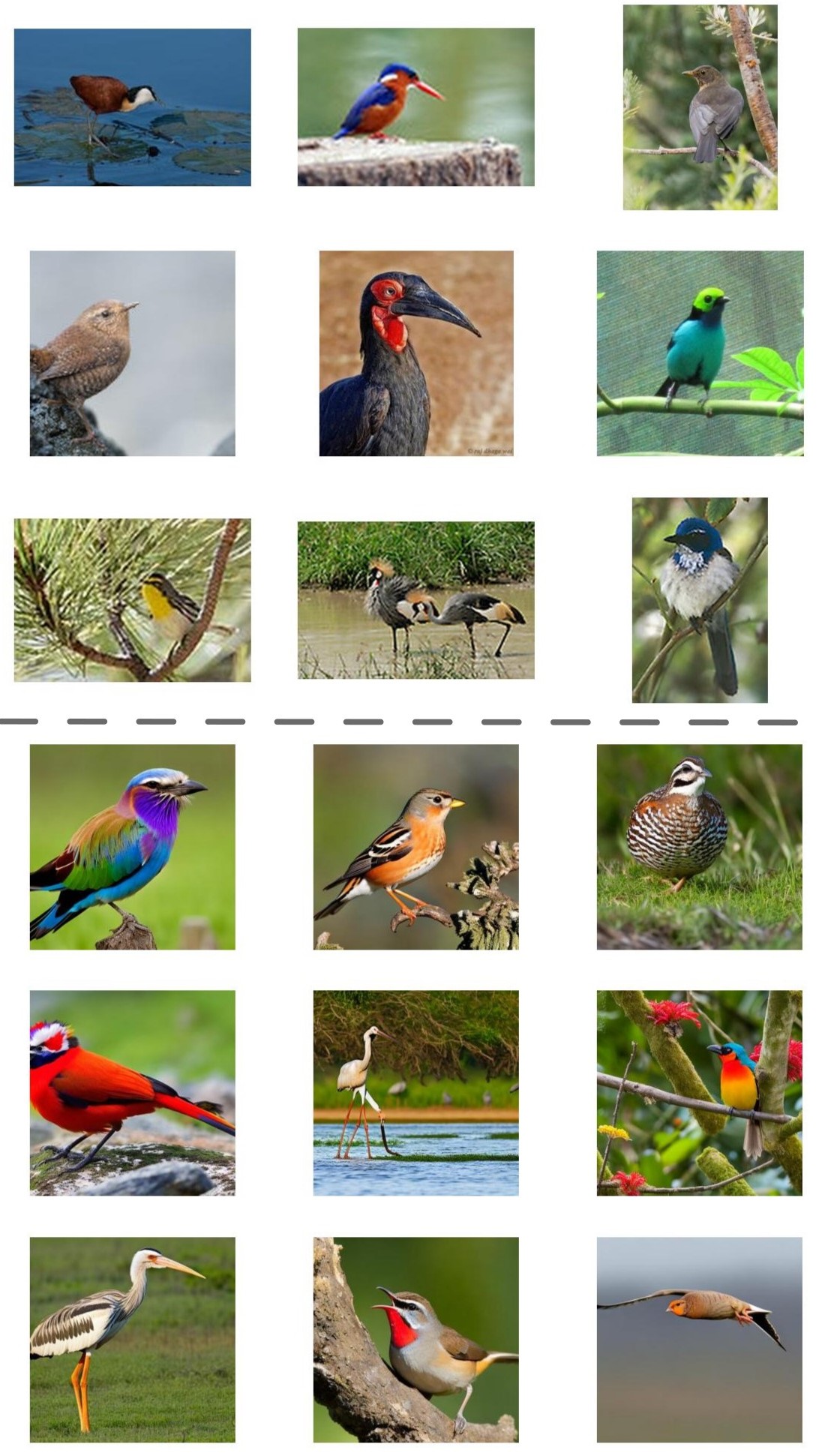}}}%
    \hfill
    \subfloat[PaS-I]{{\includegraphics[height=8.28cm]{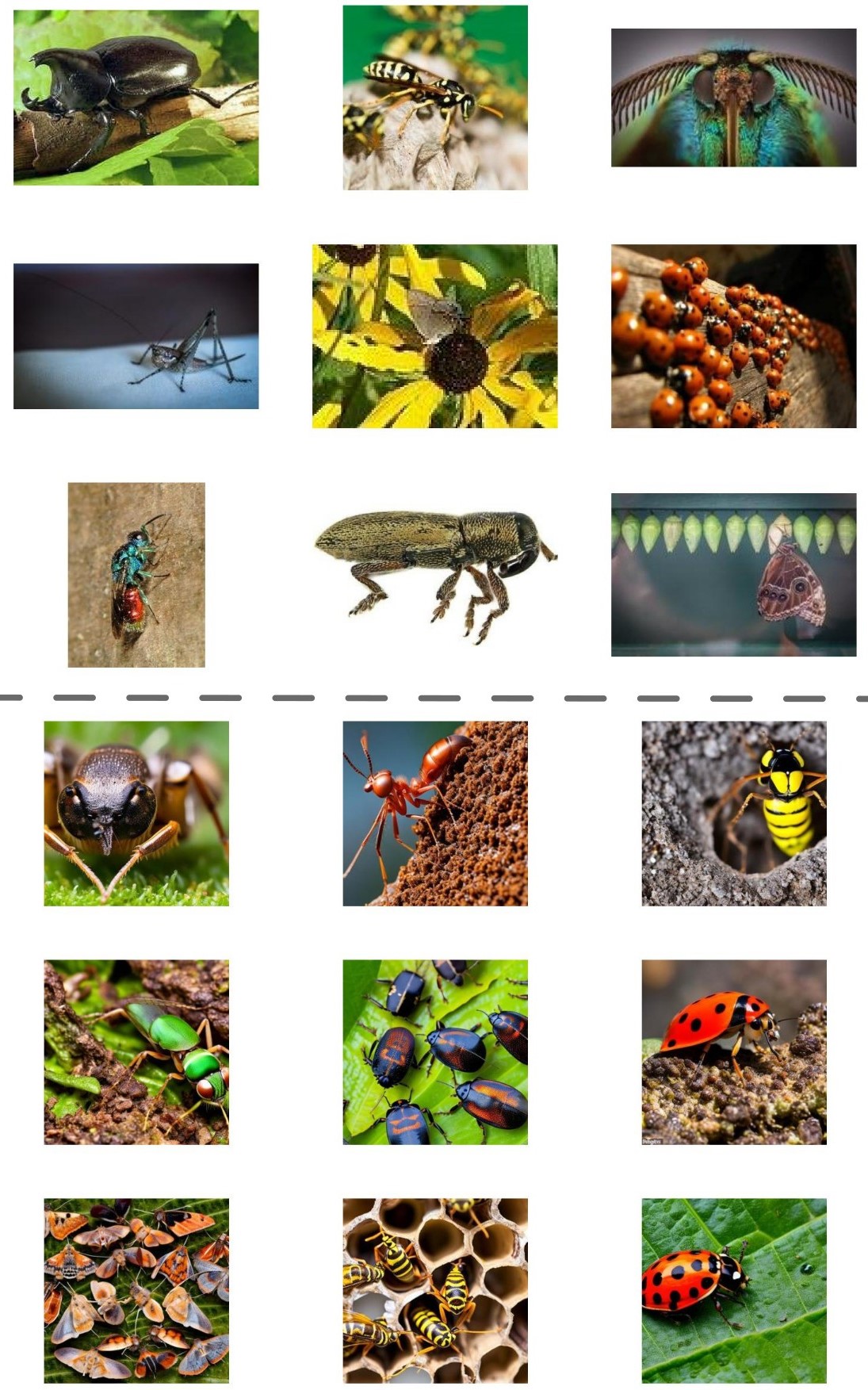}}}%
    \hfill}
    \caption{Examples of images in the final PaS datasets. For each domain, the first three rows display web images, followed by three rows of synthetic images.}
    \label{fig:pas_examples_final}
\end{figure*}

In this section, we show sample images from PaS-F, PaS-B, and PaS-I datasets during different stages of PaS. 
\newtext{The deduplication step has a significant impact, removing over 70,000, 41,000, and 27,000 near-duplicate groups for the food, birds, and insects domains, respectively.}
\Cref{fig:deduplication_examples} illustrates groups of duplicate images identified by the PaS curation pipeline across different domains. Our pipeline detects three types of duplicates: 
(1) exact duplicates, 
(2) different crops of the same image, and 
(3) images containing very similar scenes.
Having removed duplicate images, the next step in Stage 3 involves Pareto-based removal to further refine the dataset. 
\Cref{fig:pareto_examples} showcases examples from various Pareto fronts ($\mathcal{P}_1, \mathcal{P}_2, \ldots$) identified in $\mathcal{I}_{dedup}$ for each considered domain. The upper rows display images with the highest OOD scores as determined by PaS. Across all three domains, we observe a consistent pattern: there is a clear correlation between an image's Pareto position and its relevance to the target domain, underscoring the quality of the PaS datasets.
Specifically, the initial images (top rows) removed through Pareto-front based filtering are generally not closely related to the target domain. In contrast, images retained in subsequent iterations (bottom rows) exhibit greater relevance. Some retrieved web images do not align well with the target domains, highlighting the necessity of a curation step. 
\newtext{This is expected, as the underlying web-scale data pool (from Re-LAION-5B) includes varied content, from natural images to the sketches and signs visible in the figure's top rows.}
For instance, the second image in $\mathcal{P}_{1}$ of \Cref{fig:pareto_insects} demonstrates how misleading textual cues can affect the retrieval process, emphasizing the importance of the OOD metric $M_3$.
While synthetic image generation is designed to remain within the target domain—ensuring most images are well-aligned—misalignments can occur due to captioning errors. An example is the presence of houses in the first Pareto front of \Cref{fig:pareto_food}. Additionally, synthetic images that are unrealistic, such as those in the last row of \Cref{fig:pareto_insects}, are excluded from the final dataset.
Finally, \Cref{fig:pas_examples_final} contains samples from each of the final curated PaS datasets PaS-F, PaS-B, and PaS-I (9 real and 9 synthetic).

\FloatBarrier

\section{Validation of PaS Datasets}
\label{sec:results-downstream}

This section validates the effectiveness of PaS datasets as pretrainers using common SSL downstream tasks. We employ a comprehensive evaluation, including feature evaluation via k-NN and linear probing, and adaptation through full and few-shot fine-tuning. We also explore fine-tuning large VLMs using PaS image-text pairs as a way to increase their competence in each domain. 
\textcolor{black}{Additionally, we evaluate 
PaS datasets 
on real-world dense tasks in \ref{sec:appendix-other-tasks}.} 
These experiments thoroughly evaluate the quality and transferability of representations learned from PaS across visual tasks.

\subsection{Experimental setup}

\textcolor{black}{
We compare models pretrained on PaS datasets with SoTA domain-specific datasets (Food-2K, iNat\textsubscript{Birds}, and AMI-GBIF) and large-scale general domain datasets (IN-1K, IN-21K, and WIT (CLIP dataset) \cite{radford2021learning}). For all evaluations, we use a ViT-B/16 \cite{dosovitskiy2020vit}. For a fair comparison, we pretrain using a common MoCo-v3 SSL setup, except for IN-21K and WIT, for which we use official \emph{supervised} checkpoints (making them stronger discriminative baselines). To prevent data leakage, we also remove images from the PaS datasets that are similar to the evaluation test sets. The specific hyperparameters and detailed procedures are provided in \ref{sec:appendix-setup}.}

\subsection{PaS Datasets as SSL Pretrainers}

\subsubsection{Feature Evaluation with k-NN \& Linear Probing}
To ensure a fair comparison and prevent potential bias, we exclude
models overlapping with evaluation datasets (iNat\textsubscript{Birds}, Food-2K, and AMI-GBIF).
For linear probing, we use the setup detailed in \Cref{tab:finetuning_setup}, however, only training the added linear layer (the backbone remains frozen).
We report k-NN and Linear Probing performance
evaluated on common benchmarks - Food-101 \cite{bossard2014food}, FoodX-251 \cite{kaur2019foodx} for \textit{food} (\Cref{tab:cls-food}), CUB-200-2011 \cite{wah2011caltech}, and NABirds \cite{Horn_2015_CVPR} for \textit{birds} (\Cref{tab:cls-birds}) and k-NN performance on different AMI region subsets for \textit{insects} (\Cref{tab:cls-insects}).
Overall, the results highlight that models pretrained on PaS-generated datasets consistently outperform those pretrained on domain-specific datasets, with an avergae improvement of 6.7\% accross all considered tasks and domains.
In particular, even the \enquote{Mini} versions achieve higher accuracies than supervised domain-specific datasets such as iNat\textsubscript{Birds} and Food-2K (same scale), highlighting the effectiveness of our pipeline in autonomously generating high-quality domain-specific datasets without manual labeling.

\begin{table}[!t]
    \centering
    \renewcommand{\arraystretch}{0.8}
    \resizebox{\linewidth}{!}{%
    \begin{tabular}[t]{lcccccc}
        \toprule
        \multirow{2}{*}{PT Data} & \multirow{2}{*}{Size (M)} & \multicolumn{2}{c}{Food-101} & \multicolumn{2}{c}{FoodX251} & \multicolumn{1}{c}{Food-2K} \\
        &  & k-NN & Linear & k-NN & Linear & k-NN \\
        \midrule
        IN-21K$^\dag$ & 14 & 59.3 & 80.6 & 40.1 & 62.5 & 54.1 \\
        WIT$^\dag$ & 400 & 81.9 & 90.8 & 61.5 & 73.3 & 61.3 \\
        \midrule
        IN-1K & 1.2 & 56.4 & 74.1 & 43.2 & 56.2 & 57.2 \\
        Food-2K & 0.6 & 65.4 & 78.7 & 50.6 & 61.0 & - \\
        PaS-F\textsubscript{Mini} & 0.6 & \underline{66.8} & \underline{79.7} & \underline{53.7} & \underline{64.3} & \underline{59.2} \\
        PaS-F & 1.2 & \textbf{77.3} & \textbf{86.8} & \textbf{61.9} & \textbf{71.6} & \textbf{64.2} \\
        \bottomrule
    \end{tabular}%
    }
    \caption{\textbf{Classification Acc.} (\%) results for \textit{food} domain. Best (bold), Second best (underlined) comparing only SSL models. $^\dag$ represents supervised pretraining.}
    \label{tab:cls-food}
\end{table}

\begin{table}[!t]
    \centering
    \renewcommand{\arraystretch}{0.8}
    \resizebox{\linewidth}{!}{%
    \begin{tabular}[t]{lcccccc}
        \toprule
        \multirow{2}{*}{PT Data} & \multirow{2}{*}{Size (M)} & \multicolumn{2}{c}{CUB-200} & \multicolumn{2}{c}{NABirds} & iNat\textsubscript{Birds} \\
        &  & k-NN & Linear & k-NN & Linear & k-NN \\
        \midrule
        IN-21K$^\dag$ & 14 & 69.2 & 83.5 & 57.5 & 73.2 & 32.4 \\
        WIT$^\dag$ & 400 & 63.0 & 81.2 & 50.8 & 71.5 & 31.7 \\
        \midrule
        IN-1K & 1.2 & 40.8 & 49.2 & 29.2 & 36.2 & 18.3 \\
        iNat\textsubscript{Birds} & 0.4 & 19.7 & 30.6 & 11.2 & 21.2 & - \\
        PaS-B\textsubscript{Mini} & 0.4 & 33.3 & 48.6 & 22.1 & 36.6 & 12.15 \\
        PaS-B & 1.2 & \underline{46.4} & \underline{68.9} & \underline{42.3} & \underline{56.6} & \underline{24.9} \\
        PaS-B\textsubscript{Big} & 2.4 & \textbf{55.0} & \textbf{73.5} & \textbf{49.2} & \textbf{64.8} & \textbf{27.7} \\
        \bottomrule
    \end{tabular}%
    }
    \caption{\textbf{Classification Acc.} (\%) results for \textit{birds} domain. Best (bold), Second best (underlined) comparing only SSL models. $^\dag$ represents supervised pretraining.}
    \label{tab:cls-birds}
\end{table}

\begin{table}[!t]
    \centering
    \renewcommand{\arraystretch}{0.8}
    \resizebox{\linewidth}{!}{%
    \begin{tabular}[t]{lccccc}
    \toprule
\multirow{2}{*}{PT Data} & \multicolumn{4}{c}{Fine-grained (Regions)} &  \multirow{2}{*}{Binary}\\
                         & \multicolumn{1}{c}{C-America} & \multicolumn{1}{c}{N-America} & \multicolumn{1}{c}{Europe} & \multicolumn{1}{c}{All} & \\
\midrule
    IN-21K$^\dag$ & 15.1                 & 38.0                   & 34.2                 & 38.5                        & 67.9                \\
    WIT$^\dag$   & 15.1                 & 26.7   & 23.5 & 26.6        & 60.6           \\
    \midrule
    IN-1K  & \underline{16.0}                 & \underline{38.2}       & \underline{34.3}                 & \textbf{39.9}                 & 62.4                          \\
    AMI GBIF   & \multicolumn{1}{c}{-} & \multicolumn{1}{c}{-} & \multicolumn{1}{c}{-} & \multicolumn{1}{c}{-}       & \textbf{75.4}          \\
    PaS-I & \textbf{19.2}         & \textbf{38.4}         & \textbf{35.1}         & \underline{39.5}            & \underline{68.9}                           \\ \bottomrule
    \end{tabular}%
    }
    \caption{\textbf{Classification Acc.} (\%) k-NN results for \textit{insect} domain in different AMI partittions. Best (bold), Second best (underlined) comparing only SSL models. $^\dag$ represents supervised pretraining.}
    \label{tab:cls-insects}
\end{table}

PaS-F datasets exhibit remarkable performance gains (\Cref{tab:cls-food}). 
PaS-F (1.2M images) achieves a k-NN accuracy of 77.3\% on Food-101, significantly exceeding IN-1K (56.4\%) and even outperforming IN-21K (59.3\%), which is 12$\times$ larger. 
PaS-F competes closely with WIT (400M supervised dataset), outperforming it in k-NN for FoodX-251 and Food-2K.
Similarly,
PaS-B datasets (\Cref{tab:cls-birds}) consistently outperform the supervised iNat\textsubscript{Birds} on all evaluations. 
PaS-B significantly outperforms IN-1K
with improvements ranging from 5.6\% to 20.4\%. 
Although IN-21K (14M images) achieves higher accuracy, PaS-B\textsubscript{Big} (an enlarged version of PaS-B) narrows the gap from an average of 15.34\% to 9.12\% despite being nearly 6$\times$ smaller.
This proves the ability of PaS to effectively scale the size of the dataset. 
PaS-I (\Cref{tab:cls-insects}) achieves the best results in the regional subsets (Central America, North America and Europe) of the AMI dataset at the family level, outperforming both IN-1K, IN-21K and WIT. 
The improvement is highest in CA, with a relative improvement of 20\% over the second-best pretrainer.
Although IN-1K slightly outperforms PaS-I in the global task (\enquote{All}), PaS-I remains competitive with a difference below 0.4\%. 
In the \enquote{AMI Binary} classification task, PaS-I falls only behind the laboriously compiled AMI dataset (on the same scale), which was expected since the task is also part of the AMI benchmark. 
This alignment between the training (AMI) and test data is in accordance with \Cref{sec:semantic-richness-analysis}, where we identified a dominant pattern in the AMI GBIF dataset (present in all the partitions).

\subsubsection{Full fine-tuning}
For full fine-tuning, we follow the setup in \Cref{tab:finetuning_setup}. 
Given the number of experiments and the computational demands of this downstream task, the largest datasets are disregarded from this experiments. 
This includes Food-2K, iNat\textsubscript{Birds} and all the AMI partitions. 
To provide comprehensive results, the latter are included in the few-shot study (Following section).
The results are displayed in \Cref{tab:finetuning}. On the \textit{food} domain, PaS datasets exhibit better performance than existing manually curated datasets (both general and domain-specific). Particularly, PaS-F and PaS-F\textsubscript{Mini} attain the best and second-best results in both datasets, respectively. 
The larger version, PaS-F, beats the best non-PaS alternative by 2.08\% in Food-101 and 2.68\% in FoodX-251.
PaS-B outperforms both domain-specific and general datasets in CUB-200 and NABirds, with an increase of 1.63\% and 3.94\% over the best non-PaS dataset, respectively. 
PaS-B\textsubscript{Big}, further improves performance, reaching 83.2\% in CUB-200 and 77.2\% in NABirds. Furthermore, PaS-B\textsubscript{Mini}
beats iNat\textsubscript{Birds} (same scale) by at least 2.75\%.

\begin{figure*}[!t]
    \centering
    \renewcommand{\arraystretch}{0.8}
    \begin{minipage}[t]{0.44\textwidth}
   \centering
    \begin{tabular}{lccc}
    \hline
    PT                                         & Size & CUB  & NABirds \\ \hline
    IN-1K                                      & 1.2  & 78.5 & 73.2    \\
    IN-21K$^\dag$                              & 14.0 & 81.3 & 73.1    \\ \hline
    iNat\textsubscript{Birds}                  & 0.4  & 76.8 & 70.0    \\
    PaS-B\textsubscript{Mini}                  & 0.4  & 79.5 & 73.0    \\
    PaS-B                                      & 1.2  & \underline{82.9} & \underline{77.1}    \\
    PaS-B\textsubscript{Big}                   & 2.4  & \textbf{83.2} & \textbf{77.2}    \\ \hline
    \end{tabular}
    \end{minipage}%
    \hfill
    \begin{minipage}[t]{0.42\textwidth}
   \centering
    \vspace{-4.6em}
    \begin{tabular}{lccc}
    \hline
    PT                                         & Size & F101 & FX251 \\ \hline
    IN-1K                                      & 1.2  & 87.1 & 71.7  \\
    IN-21K$^\dag$                              & 14.0 & 86.3 & 70.1  \\ \hline
    Food-2K                                    & 0.6  & 87.9 & 72.5  \\
    PaS-F\textsubscript{Mini}                  & 0.6  & \underline{88.3} & \underline{73.6}  \\
    PaS-F                                      & 1.2  & \textbf{89.1} & \textbf{74.3} \\ \hline
    \end{tabular}
    \end{minipage}
     \captionof{table}{
     \textbf{Full fine-tuning} 
     on \textit{food} and \textit{birds}.  In all the cases, the architecture used is ViT-B/16. 
     $\dag$ - supervised denotes backbones pretrained in a supervised way.
     }\label{tab:finetuning}
\end{figure*}

\subsubsection{Few-shot fine-tuning.}

\begin{figure*}[!t]
    \centering
    \includegraphics[width=\textwidth]{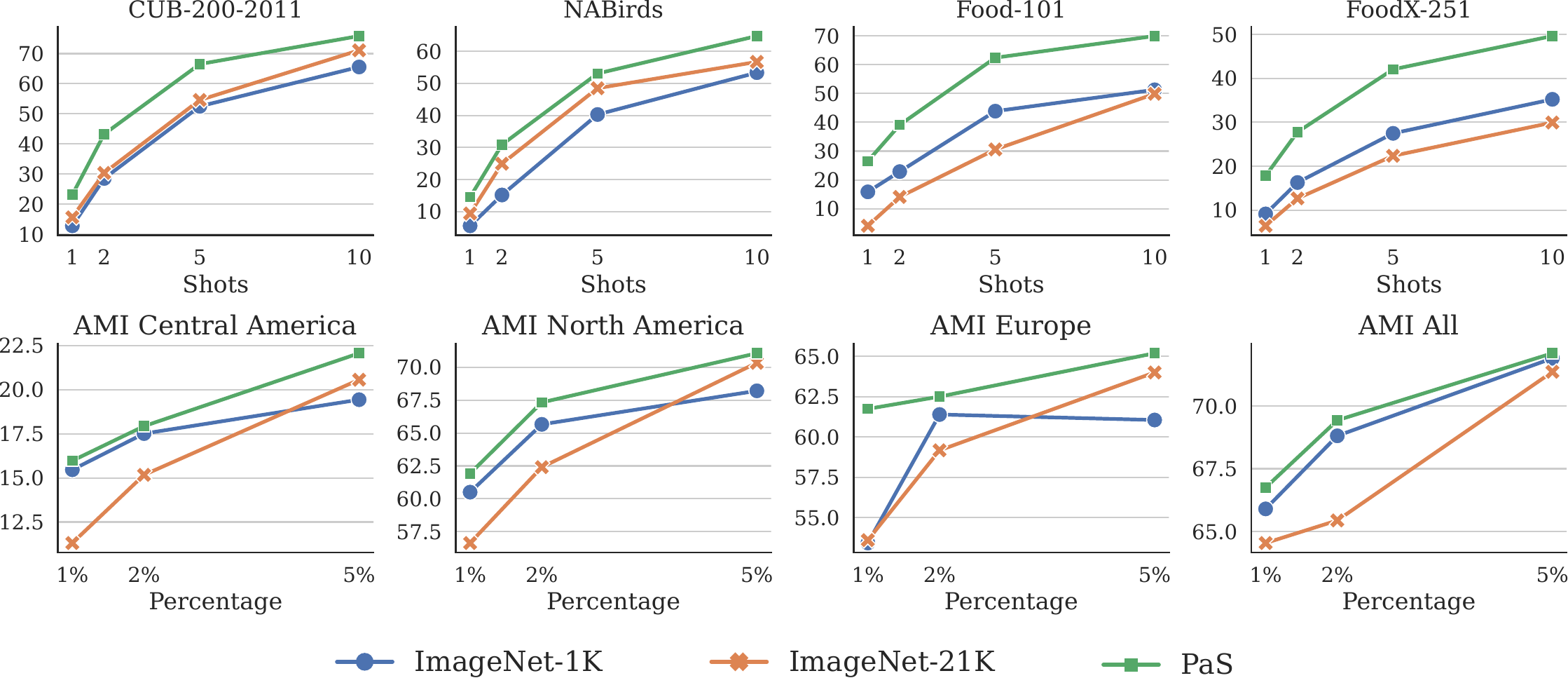}
    \caption{\textbf{Results on few-shot fine-tuning} on ViT-B/16.}\label{fig:few_shot}
\end{figure*}

The ability to achieve high performance with limited, labeled data is crucial in many real-world applications, where acquiring large annotated datasets is expensive or impractical. 
Few-shot learning addresses this challenge by aiming to learn effectively from only a few examples per class. To evaluate the data efficiency of representations learned from PaS datasets, we perform fine-tuning with reduced portions of training datasets in the three domains (results displayed in \Cref{fig:few_shot}). Backbones pretrained on PaS datasets outperform general-domain datasets in all considered scenarios, while being even an order of magnitude smaller.
PaS-pretrained backbones achieve competitive performance (even surpass) other backbones when using fewer images for fine-tuning, highlighting, the potential of PaS datasets to be data-efficient and show better learnability of useful features for the desired domain. 
For this downstream task, we use the same setup as full fine-tuning, only reducing the number of samples per class.

\subsection{Fine-tuning of VLMs} \label{sec:results-vlm}

Multimodality is a core trait of our PaS pipeline. As a result, the final generated image dataset is a paired text associated with each image (the web caption in the case of web-retrieved images, and the generation prompt in the case of synthetic images). 
To assess the capacity and relevance of PaS in this regard, we leverage the text-image pairs
to fine-tune pretrained VLMs, namely CLIP \cite{radford2021learning} and SigLIP \cite{zhai2023sigmoid}, using LoRA \cite{hu2022lora}.
In this task, the original weights of the VLM remain frozen, and only the LoRA matrices are updated. 
Specifically, LoRA is applied to the Q, K, and V linear layers of all transformers in the vision encoder. 
For both CLIP \cite{radford2021learning} and SigLIP \cite{zhai2023sigmoid} fine-tuning, we adopt the same hyperparameters as in \cite{zanella2024low}, with a scheduler of 5 epochs. This approach allows for efficient fine-tuning by less than only 1\% of the model parameters using PaS' image-text pairs. 

For evaluation, we take the visual encoder of the fine-tuned model and use it for k-NN and linear evaluation. 
We use
ViT-B/16-based CLIP and SigLIP,
pretrained on massive web-scale data: 400M and 4.8B images, respectively. 
Results in \Cref{tab:VLFT} show that PaS leverages the knowledge of CLIP to create datasets with enriched knowledge, yielding incremental improvements across domains with an average gain of 4.2\%. Regarding SigLIP, PaS datasets enable improvements in all domains while only representing a small fraction (0.03\%) of the original pretraining data of SigLIP. The results highlight the effectiveness of PaS in creating targeted fine-tuning datasets for complex domains, enabling improvements even when retraining foundational models is impractical due to continuous new data and domains.

\begin{table}[t]
	\centering
    \renewcommand{\arraystretch}{0.8}
	\resizebox{\linewidth}{!}{%
	\begin{tabular}{lccccccccc}
		\hline
		\multicolumn{1}{c}{} & \multirow{2}{*}{\textbf{\#Imgs}} & \multicolumn{2}{c}{\textbf{CUB}} & \textbf{iNat\textsubscript{B}} & \multicolumn{2}{c}{\textbf{F101}} & \multicolumn{1}{l}{\textbf{F2K}} & \textbf{AMI\textsubscript{CA}} & \textbf{AMI\textsubscript{All}}   \\ \cline{3-10}
		\multicolumn{1}{c}{}                       &             & \textbf{kNN}  & \textbf{Lin.}  & \textbf{kNN}  & \textbf{kNN}  & \textbf{Lin.}  & \textbf{kNN}   & \textbf{kNN} & \textbf{kNN}    \\ \hline
		\textbf{CLIP} {\tiny (WIT)}                        & 400M        & 63.0          & 81.2           & 61.3          & 81.9          & 90.8           & 31.7           & 15.1 &   26.6         \\
		\textbf{SigLIP}                            & $\sim$ 4.8B & 70.8          & 82.6           & 70.0          & 89.7          & 93.9           & 35.5           & 16.5 &   39.0        \\ \hline
		\textbf{PaS}                               & 1.2M        & 46.4          & 68.9           & 64.2          & 77.3          & 86.8           & 24.9           & \textbf{19.2} &  \textbf{39.5} \\
		\textbf{\hspace{0.5em}\& CLIP$^\dagger$}   & +1.2M       & 67.4          & 82.8           & 67.6          & 87.3          & 92.2           & \textbf{37.5}  & 15.9  &    36.1      \\
		\textbf{\hspace{0.5em}\& SigLIP$^\dagger$} & +1.2M       & \textbf{71.3} & \textbf{82.9}  & \textbf{70.7} & \textbf{90.1} & \textbf{94.2}  & 35.9           & 16.7   &    39.4     \\ \hline
	\end{tabular}%
	}
	\caption{PaS vs. CLIP and SigLIP using ViT-B/16 as visual encoder. We fine-tuned the VL models using LoRA with PaS datasets for 5 epochs.} \label{tab:VLFT}
\end{table}

\subsection{Ablations}

We perform ablations
of key components of PaS
(\Cref{tab:ablations}) using the \textit{food} domain.
\begin{enumerate}[wide = 0pt, label={(\arabic*)}]
    \item \textbf{Concept Discovery:} 
    Our LLM-guided concept generation outperforms a PaS dataset curated directly from Food-101 concepts, highlighting greater generalization and diversity.  Evaluating Stage 1, our LLM-based concept generation (\textbf{PaS-F$_{\text{Mini}}$}) outperforms a manually curated set of concepts from Food-101 ( \textbf{PaS-F$_{F101}$} ), indicating greater generalization and diversity.
    \item \textbf{Data Distribution:}
    We ablate the automatic \newtext{synthetic-to-real} ratio of PaS by manually setting them.
    \newtext{To achieve these precise ratios for the ablation, the curation methodology was modified: while deduplication is performed on the union of real and synthetic data to remove all near-duplicates globally, the subsequent Pareto-front filtering is applied to the real and synthetic sets independently.}
    Comparing manually set ratios (\newtext{\textbf{PaS-F$_{\text{Mini}}^{0-100}$}}, \textbf{PaS-F$_{\text{Mini}}^{25-75}$},  \textbf{PaS-F$_{\text{Mini}}^{50-50}$},\torm{and}\textbf{PaS-F$_{\text{Mini}}^{75-25}$}\newtext{, and \textbf{PaS-F$_{\text{Mini}}^{100-0}$}}) with the automatically distribution determined by PaS, 
    \textbf{PaS-F\textsubscript{Mini}} achieves competitive (a close second)
    performance without manual tuning, demonstrating an effective balance between both data sources. \newtext{More importantly, while both pure synthetic and pure web datasets are effective, their hybrid combination consistently yields the best performance, reinforcing the value of our hybrid data collection approach.}
    \item \textbf{Data Curation:} 
    We study the effectiveness of our dataset reduction by comparing unfiltered (\textbf{PaS$_{NF}$}), deduplicated (\textbf{PaS$_{SF}$}) and Pareto front-based OOD removal (\textbf{PaS-F}) of which the complete reduction yields the best performance, with fewer but more relevant images.
\end{enumerate}

In summary, our ablation experiments highlight the effectiveness of each component of the PaS pipeline. The LLM-guided concept generation provides superior generalization, the automatic web-to-synthetic ratio achieves a balanced performance without manual tuning, and the data curation process effectively removes irrelevant images, leading to the best overall results.

\addtolength{\tabcolsep}{-0.3em}
\begin{table*}[!t]
    \centering
    \renewcommand{\arraystretch}{0.8}
\begin{tabular}{lcccccccccc}
\toprule
PT Data                                                          & LLM & Synth & Web & SF & PF & \multicolumn{2}{c}{Food-101} & \multicolumn{2}{c}{FoodX-251} & \multicolumn{1}{c}{F-2K}   \\ \cmidrule(lr){7-8} \cmidrule(lr){9-10} \cmidrule(lr){11-11}
&&&&&& $k$-NN & Linear & $k$-NN & Linear & $k$-NN  \\ \midrule
 PaS-F$_{F101}$                                                       &      & A  & A & \checkmark           & \checkmark            & 67.03 & 81.12 & 52.94 & 64.12 & 59.22  \\
 \newtext{PaS-F$_{\text{Mini}}^{0-100}$} & \newtext{\checkmark}    & \newtext{0}    & \newtext{100}    & \newtext{\checkmark}           & \newtext{\checkmark}            & \newtext{65.79} & \newtext{79.20} & \newtext{52.80} & \newtext{63.87} & \newtext{59.24}  \\
  PaS-F$_{\text{Mini}}^{25-75}$ & \checkmark    & 25    & 75   & \checkmark           & \checkmark            & 65.84 & 79.11 & 52.83 & 63.84 & 59.30  \\
 PaS-F$_{\text{Mini}}^{50-50}$                                        & \checkmark   & 50    & 50   & \checkmark           & \checkmark     &  66.04 & 79.38 & 53.07 & 63.98 & 59.83  \\
 PaS-F$_{\text{Mini}}^{75-25}$ & \checkmark  & 75    & 25   & \checkmark           & \checkmark            & 66.79 & 79.93 & 53.82 & 64.38 & 60.67  \\
  \newtext{PaS-F$_{\text{Mini}}^{100-0}$} & \newtext{\checkmark}    & \newtext{100}     & \newtext{0}    & \newtext{\checkmark}           & \newtext{\checkmark}            & \newtext{66.53} & \newtext{79.42} & \newtext{53.64} &  \newtext{64.12}  &  \newtext{60.05}   \\
 PaS-F$_{\text{Mini}}$             & \checkmark    & A  & A & \checkmark           & \checkmark            & 66.75 & 79.68 & 53.68 & 64.29 & 60.46  \\\midrule
 PaS$_{NF}$                                                     & \checkmark   & A  & A &             &             & 70.89 & 81.50 & 57.39 & 66.33 & 61.23  \\
 PaS$_{SF}$                                                    & \checkmark  & A  & A & \checkmark           &      & 73.11 & 83.48 & 59.44 & 68.15 & 62.45  \\
 PaS-F    & \checkmark  & A  & A & \checkmark           & \checkmark      &   \textbf{77.31} & \textbf{86.84} &   \textbf{61.86} & \textbf{71.61} &   \textbf{64.23}  \\
\bottomrule
\end{tabular}
\captionof{table}{
\textbf{Ablation} on \textit{food} using ViT-B/16 ($k$-NN and Linear probing). Columns: LLM-based concept discovery (LLM), Synth/Web proportions ('A'=automatic), deduplication (SF), and Pareto filtering (PF). Evaluation results are reported for Food-101, FoodX-251, and Food-2K on two tasks ($k$-NN and Linear).
}
\label{tab:ablations}
\end{table*}
\addtolength{\tabcolsep}{+0.3em}

\subsection{Limitations}
While PaS generates high-quality datasets and demonstrates promising pretraining results, it is crucial to identify its limitations.
\begin{enumerate}[wide = 0pt, label={(\arabic*)}]
\item
While PaS leverages the knowledge of large pretrained models, such as LLMs and text-to-image generators, its effectiveness can be influenced by their general capabilities. 
Even if the design of PaS has been proven robust in several complex domains, limitations or biases in these foundational models for highly specialized domains might affect the generated dataset's applicability.

\item
Evaluation pipeline is similarly sensitive, as the quality of the target domain can impact latent representations and lexical alignment analysis. However, PaS's modular architecture allows for easy replacement or upgrading of these components, effectively mitigating both limitations. 
\item
PaS leverages efficient large models, however, depending on model choice, computational demand may increase, potentially limiting the quality in low-resource settings. 
Using techniques such as model quantization can alleviate this limitation.
\end{enumerate}

\section{Conclusion}\label{sec:conclusion}

While being critical for real-world deployment, creating domain-specific datasets at scale is challenging due to its high costs.
In this paper, we introduce PaS, an innovative framework for autonomously generating domain-specific datasets on-demand. 
PaS is modular, facilitating integration of various SoTA components like LLMs and VLMs, offers adaptability across diverse domains, and, allows future advancements.
PaS incorporates efficient pruning techniques to provide high-performance relevant datasets with reduced size, tackling a key challenge in dataset curation.
Our comprehensive task-agnostic analysis
highlights our framework's ability to produce scalable datasets that surpass the richness and diversity of conventionally curated domain-specific SoTA datasets. 
When pretrained on PaS datasets, models achieve competitive or superior results compared to traditional supervised datasets of the same scale.  
Empirical results show that PaS datasets outperform IN-1K across all tested domains and even surpass IN-21K supervised setup in \textit{food} and \textit{insects} domain while being an order of magnitude smaller.
Moreover, PaS multi-modality enables cost-effective adaption of existing large VLMs to new domains, easing their usage in new applications.
PaS represents a paradigm shift:
automatically generated domain-specific datasets can bridge the gap with large general-domain datasets, providing scalable, flexible solutions with enhanced performance in specialized applications.
The efficiency of PaS makes it a promising approach for real-world deployment, including resource-constrained environments.
Future directions include testing advanced components for complex domains, analysing data and distribution biases, and training cross-modal models.

\section*{Acknowledgments}

\hyphenation{Deep-Sense}
\newcommand{\museNo}{01070421}

\newcommand{\DFVolNo}{PDC\-2022-133642-I00}

This work was partially funded by the EU project MUSAE (No. \museNo), 2021-SGR-01094 (AGAUR), Icrea Academia'2022 (Generalitat de Catalunya), Robo STEAM (2022-1-BG01-KA220-VET-000089434, Erasmus+ EU), DeepSense (ACE053/22/000029, ACCIÓ), DeepFoodVol (AEI-MICINN, \DFVolNo) and IDEATE (AEI-MICINN, PID2022-141566NB-I00).
J. M. Rodríguez-de-Vera and Imanol G. Estepa acknowledge the support of FPU Becas with code FPU22/03116 and FPU23/02822 respectively, Ministry of Universities, Spain.
B. Nagarajan acknowledges AI4S fellowship within the “Generación D” initiative by Red.es, Ministerio para la Transformación Digital y de la Función Pública, for talent attraction (C005/24-ED CV1), funded by NextGenerationEU through PRTR.
The authors thankfully acknowledge 
EuroHPC Joint Undertaking (EHPC-DEV-2023D12-059) for awarding us access to Leonardo at CINECA, Italy and Spanish Supercomputing Network (RES) (IM-2023-3-0019) for awarding us access to MareNostrum5 at BSC, Spain.

\FloatBarrier

\bibliographystyle{unsrtnat}
\bibliography{references}

\clearpage

\appendix

\section{PaS Dataset Diversity \& Domain Assessment - Semantic richness analysis in the insects domain}\label{sec:appendix-insects}

\begin{figure}[!htpb]
    \centering
    \vspace{0.1em}
    \includegraphics[width=0.45\textwidth, angle=-90]{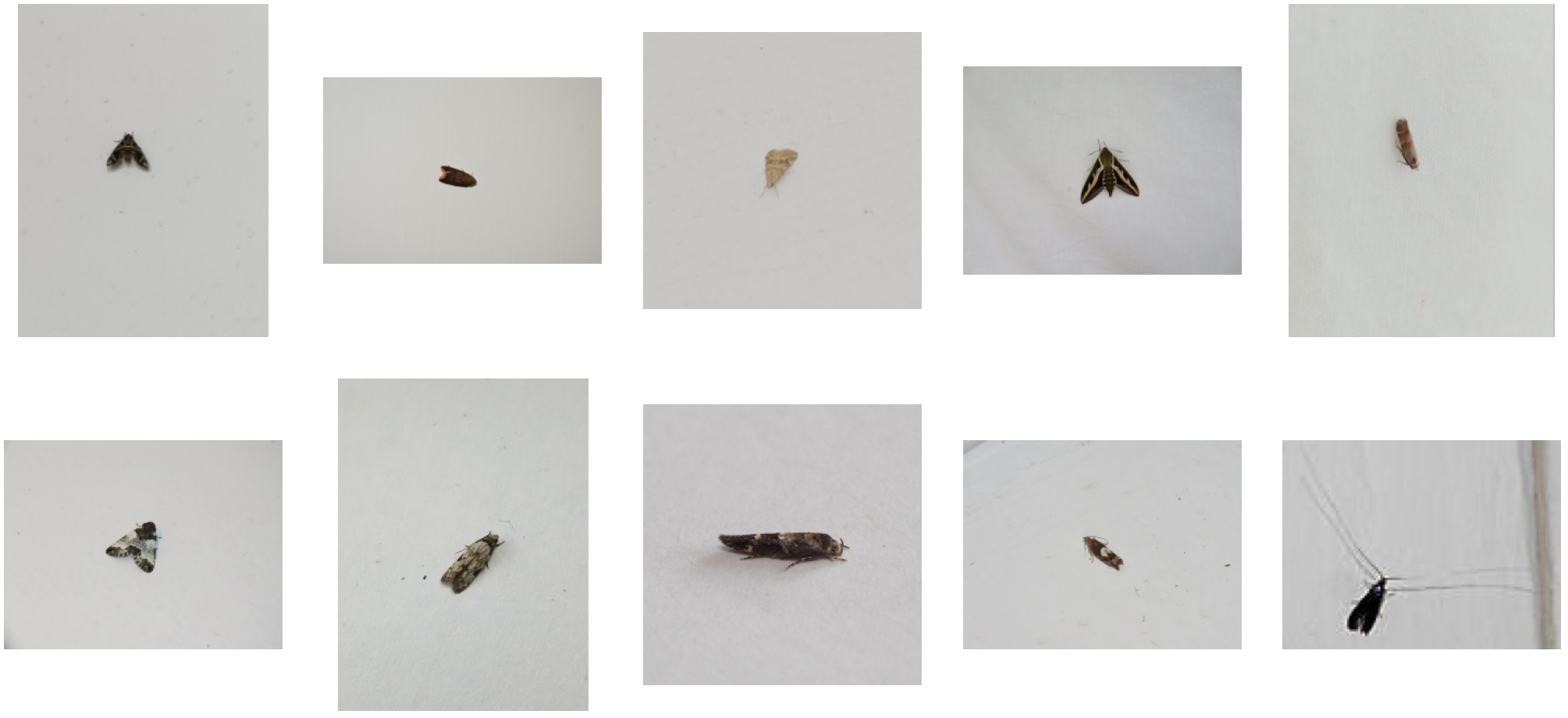}
    \caption{Samples from the AMI GBIF dataset belonging to its most populated prototype.}\label{fig:proto-ami}
\end{figure}

\textcolor{black}{As mentioned in \Cref{sec:semantic-richness-analysis}, the most populated AMI-GBIF prototype represents a human-driven concept (quantity over diversity).} 
PaS-I,
manages to include similar concept images with much fewer samples, as it focuses on the \textit{insects} domain without a human strategy that could bias the dataset.
Specifically, this prototype appears in 38.94\% of AMI GBIF images, compared to only 0.89\% in PaS-I.
\Cref{fig:proto-ami} showcases examples of images associated with this prototype. These images predominantly feature well-centered insects (mainly moths) captured on white surfaces. In contrast, PaS-I images—both web-sourced and synthetic—typically depict insects in natural environments. This fundamental difference accounts for the distribution discrepancies highlighted in the previous section.
The prototypes across different domains demonstrate the richness of PaS datasets, highlighting that PaS datasets
not only match but potentially exceed the semantic diversity of existing human-curated datasets.

\section{Additional evaluations of PaS datasets as pretrainers}\label{sec:appendix-other-tasks}

\subsection{Domain-Specific Evaluation.} 
\textcolor{black}{In addition to the comprehensive experimentation in \Cref{sec:results-downstream},} we assess our pretrained models on two specialized real-world tasks:
semantic segmentation for \textit{food} using FoodSeg103 dataset \cite{wu2021large} and keypoint detection for \textit{birds} using Birdsnap dataset \cite{berg2014birdsnap}.

\paragraph{Semantic Segmentation of \textit{Food}}

\Cref{tab:segmentation} contains the results for \textit{food} semantic segmentation using \textit{FoodSeg103} \cite{wu2021large}. 
In all cases, we use default configurations provided by MMSegmentation\footnote{\url{https://github.com/open-mmlab/mmsegmentation}} without any hyper-parameter tuning. 
We use a scheduler of 80K training steps.
The results indicate that, ViT-B backbone pretrained with PaS-F outperforms other backbones, including both ViT-B and Swin \cite{liu2021swin}, trained on different datasets. 
Notably, significant improvements are observed in both mIoU and mAcc metrics. These findings
highlight the effectiveness of specialized datasets like PaS-F in enhancing performance on dense prediction tasks such as semantic segmentation.

\begin{table}[!htpb]
\centering
\small
\renewcommand{\arraystretch}{0.8}
\begin{tabular}{@{}lclrr@{}}
\toprule
\multicolumn{1}{l}{PT Data}  & Method                & Backbone       & \multicolumn{1}{c}{mIoU} & \multicolumn{1}{c}{mAcc} \\ \midrule
\multirow{3}{*}{IN-21K} & \multirow{1}{*}{SETR \citep{SETR}} & ViT-B/16       & 41.3                     & 52.7                     \\ \cmidrule(l){2-5}  
                              & \multirow{2}{*}{UperNet \citep{xiao2018upernet}}  & Swin-S       & \underline{41.6}                     & 53.6                     \\
                              &                        & Swin-B         & 41.2                     & \underline{53.9}                     \\ \midrule
Food-2K                       & \multirow{2}{*}{UperNet \citep{xiao2018upernet}}  & ViT-B/16       & 38.1                     &  51.1                    \\
PaS-F                         &                        &  ViT-B/16       & \textbf{42.5}                     & \textbf{55.5}                     \\ \bottomrule
\end{tabular}
\caption{
\textbf{Results on the downstream task of semantic segmentation for the dataset {\em FoodSeg103} \cite{wu2021large}.} 
Results for {\em ImageNet-21K} pretraining have been taken from the {\em FoodSeg103} paper \cite{wu2021large}.
The other experiments follow the same setup.}
\label{tab:segmentation}
\end{table}

\paragraph{Keypoint Detection on \textit{Birds}}

For keypoint detection using Birdsnap \cite{berg2014birdsnap}, we compare
ViT-B/16 performance pretrained on IN-21K with that
pretrained on PaS-B\textsubscript{Big}, using ViTPose \citep{xu2022vitpose} with default hyperparameters. 
Despite PaS-B\textsubscript{Big} being an order of magnitude smaller, it achieves superior results in mean mAP: 59.6\% compared to 56.3\% for IN-21K pretrained models.

\section{Experimental Setup}\label{sec:appendix-setup}

\textcolor{black}{For all our experiments, we use ViT-B/16 as the visual encoder \cite{dosovitskiy2020vit}. To ensure a fair comparison across datasets, we pretrain the encoders from scratch using a common SSL setup: MoCo-v3 \cite{chen2021empirical} for 300 epochs. The only exceptions are IN-21K and WIT, for which we use the officially released, supervisedly trained weights \cite{radford2021learning,ridnik2021imagenet21k}, making them stronger discriminative baselines.}

\textcolor{black}{To ensure that our evaluation is not biased by overlapping data, we meticulously remove images from the PaS datasets that are similar to those in the downstream evaluation test sets. For this, we use the state-of-the-art duplicate detection model, SSCD \cite{pizzi2022self}, following the procedure outlined in \cite{oquab2024dinov2}. This process is analogous to the deduplication step in our curation pipeline (see \Cref{sec:dataset_curation}) but employs a stricter similarity threshold of 0.45 to aggressively remove any potential near-duplicates of the test images from our final pretraining datasets.}

\begin{table}[!htpb]
    \centering
    \renewcommand{\arraystretch}{0.8}
    \begin{subtable}[b]{0.45\textwidth}
        \centering
        \begin{tabular}{@{}lr@{}}
            \toprule
            Parameter             & Value \\
            \midrule
            Backbone             & VIT-B/16 \\
            Batch Size           & 4096 \\
            Learning Rate        & $2.4 \times 10^{-3}$ \\
            Scheduler & Cosine \\
            Optimizer            & AdamW \\
            Epochs      & 300 \\
            Warmup Period        & 40 \\
            Weight Decay         & 0.1 \\
            \bottomrule
        \end{tabular}
        \caption{Moco v3 pretraining setup}
        \label{tab:mocov3_pretrain}
    \end{subtable}
    \hfill
    \begin{subtable}[b]{0.45\textwidth}
        \centering
        \begin{tabular}{@{}lr@{}}
            \toprule
            Parameter             & Value \\
            \midrule
            Backbone             & VIT-B/16 \\
            Batch Size           & 256 \\
            Learning Rate        & $5 \times 10^{-4}$ \\
            Scheduler & Cosine \\
            Optimizer            & AdamW \\
            Epochs      & 100 \\
            Warmup Period        & 10 \\
            Weight Decay         & 0.05 \\
            \bottomrule
        \end{tabular}
        \caption{Fine-tuning setup}
        \label{tab:finetuning_setup}
    \end{subtable}
    \caption{Detailed setups used for the different pretraining and fine-tuning.}
    \label{tab:comparison}
\end{table}

\end{document}